# On Ensuring that Intelligent Machines Are Well-Behaved

Philip S. Thomas,[1] Bruno Castro da Silva,[2]
Andrew G. Barto,[1] and Emma Brunskill[3]

[1]University of Massachusetts Amherst
[2]Universidade Federal do Rio Grande do Sul
[3]Stanford University

Machine learning algorithms are everywhere, ranging from simple data analysis and pattern recognition tools used across the sciences to complex systems that achieve super-human performance on various tasks. Ensuring that they are well-behaved—that they do not, for example, cause harm to humans or act in a racist or sexist way—is therefore not a hypothetical problem to be dealt with in the future, but a pressing one that we address here. We propose a new framework for designing machine learning algorithms that simplifies the problem of specifying and regulating undesirable behaviors. To show the viability of this new framework, we use it to create new machine learning algorithms that preclude the sexist and harmful behaviors exhibited by standard machine learning algorithms in our experiments. Our framework for designing machine learning algorithms simplifies the safe and responsible application of machine learning.

## 1  Introduction

Machine learning algorithms are impacting our lives—they are used across the sciences, for example, by geologists to predict landslides [Jibson, 2007] and biologists working to create a vaccine for HIV [Bhasin and Raghava, 2004], they influence criminal sentencing [Angwin et al., 2016], they control autonomous vehicles [Pomerleau, 1989], and they are transforming health care delivery [Saria, 2014]. The potential for machine learning algorithms to cause harm—including catastrophic harm—is therefore a pressing concern [Bostrom, 2014]. Despite the importance of this problem, current machine learning algorithms do not provide their users with an effective means for precluding undesirable behavior, which makes the safe and responsible use of machine learning algorithms a difficult and error prone process. We introduce a new framework for designing machine learning algorithms that allow their users to easily define and regulate undesirable behaviors. This framework does not address the problem of imbuing intelligent machines with a notion of morality or human-like values [Russell, 2016], nor the problem of avoiding undesirable behaviors that the user never considered [Amodei et al., 2016]. Rather, it provides a remedy for the problem of machine learning algorithms that misbehave because their users did not have an effective way to specify and constrain undesirable behaviors.

The first step of the current standard approach for designing machine learning algorithms, which we refer to as the *standard approach* hereafter, is to define mathematically what the algorithm should do. At an abstract level, this definition is the same across all branches of machine learning: find a solution, $\theta^\star$, within a *feasible set*, $\Theta$, that maximizes an *objective function*, $f : \Theta \to \mathbb{R}$. That is, the goal of the algorithm is to find a solution in

$$\arg\max_{\theta \in \Theta} f(\theta). \tag{1}$$

Importantly, the algorithm does not know $f(\theta)$ for any $\theta \in \Theta$ (e.g., the true mean squared error)—it can only reason about it from data (e.g., by using the sample mean squared error).

One problem with the standard approach is that the user of a machine learning algorithm must encode constraints on the algorithm's behavior in the feasible set or objective function. Encoding constraints in the objective function (e.g., using soft constraints [Boyd and Vandenberghe, 2004] and robust and risk-sensitive approaches [Bertsimas et al., 2004]) requires extensive domain knowledge or additional data analysis to properly balance the relative importance of the primary objective function and the constraints. Furthermore, maximizing an objective function that appears to reasonably represent desired behavior can sometimes result in behavior that is unacceptable [von Goethe, 1797]. Similarly, encoding constraints in the feasible set (e.g., using hard constraints [Boyd and Vandenberghe, 2004], chance-constraints [Charnes and Cooper, 1959], and robust optimization approaches [Ben-Tal et al., 2009]) requires knowledge of the probability distribution from which the available data is sampled, which is often not available.

Our framework for designing machine learning algorithms allows the user to constrain the behavior of the algorithm more easily—without requiring extensive domain knowledge or additional data analysis. This is achieved by shifting the burden of ensuring that the algorithm is well-behaved from the user of the algorithm to the designer of the algorithm. This is important because machine learning algorithms are used in critical applications by people who are experts in their fields, but who may not be experts in machine learning and statistics. In general, it is impossible to guarantee that undesirable behavior will never occur when using machine learning algorithms. Algorithms designed using our new framework therefore preclude undesirable behaviors *with high probability*.

We now define our new framework. Let $D \in \mathcal{D}$ (a random variable) be the data provided as input to the machine learning algorithm, $a$, where $a : \mathcal{D} \to \Theta$, so that $a(D)$ is a random variable that represents the solution produced by the algorithm when given data $D$ as input. Our framework is based on a different way of mathematically defining what an algorithm should do, which allows the user to directly place probabilistic constraints on the solution, $a(D)$, returned by the algorithm. This differs from the standard approach wherein the user can only indirectly constrain $a(D)$ by restricting or modifying the feasible set, $\Theta$, or objective function, $f$. Concretely, algorithms constructed using our framework are designed to satisfy constraints of the form: $\Pr(g(a(D)) \leq 0) \geq 1 - \delta$, where $g : \Theta \to \mathbb{R}$ defines a measure of undesirable behavior (as illustrated later by example), and $\delta \in [0, 1]$ limits the admissible probability of undesirable behavior. Since these constraints define which algorithms, $a$, are acceptable (rather than which solutions, $\theta$, are acceptable), they must be satisfied during the design of the algorithm rather than when the algorithm is applied. This shifts the burden of ensuring that the algorithm is well-behaved from the user to the designer.

Using our framework involves three steps:

1. The designer of the algorithm writes a mathematical expression that expresses his or her goal; in particular, the properties that he or she wants the resulting algorithm, $a$, to have. This expression has the following form, which we call a *Seldonian optimization problem* after the eponymous character in Asimov's *Foundation* [Asimov, 1951]:

$$\arg\max_{a \in \mathcal{A}} f(a) \qquad (2)$$
$$\text{s.t. } \forall i \in \{1, \ldots, n\},\, \Pr(g_i(a(D)) \leq 0) \geq 1 - \delta_i,$$

   where $\mathcal{A}$ is the set of all algorithms that will be considered, $f : \mathcal{A} \to \mathbb{R}$ is now an objective function that quantifies the utility of an algorithm, and we allow for $n \geq 0$ constraints, each defined by a tuple, $(g_i, \delta_i)$, where $i \in \{1, \ldots, n\}$. Notice that this is in contrast to the standard approach—in the standard approach the mathematical expression, (1), defines the goal of the *algorithm*, which is to produce a *solution* with a given set of properties, while in our framework the mathematical expression, (2), defines the goal of the *designer*, which is to produce an *algorithm* with a given set of properties.

2. The user should have the freedom to define one or more $g_i$ that captures his or her own desired definition of undesirable behavior, and should be able to do so without knowledge of the distribution of $D$ or even the value of $g_i(\theta)$ for any $\theta \in \Theta$. This requires the algorithm, $a$, to be compatible with many different definitions of $g_i$. The designer should therefore specify the class of possible definitions of $g_i$ with which the algorithm will be compatible, and should provide a means for the user to tell the algorithm which definition of $g_i$ should be used. Later we provide examples of how this can be achieved.

3. The designer creates an algorithm, $a$, which is a (possibly approximate) solution to the mathematical statement in the first step, and which allows for the class of $g_i$ chosen in the second step. In practice, designers rarely produce algorithms that cannot be improved upon, which implies that they may only find *approximate* solutions to (2). Our framework allows for this by only requiring $a$ to satisfy the probabilistic constraints while *attempting* to optimize $f$; we call such algorithms *Seldonian*.

Once a Seldonian algorithm has been designed, it has built-in guarantees about the probability that it will produce undesirable behavior, and a user can apply it by specifying one or more $g_i$ to capture his or her desired definition of undesirable behavior, and $\delta_i$ to be the maximum admissible probability of the undesirable behavior characterized by $g_i$.

To show the viability of our framework, we used it to design regression and reinforcement learning algorithms. Constraining the behavior of regression algorithms is important because, for example, they have been used for medical applications where undesirable behavior could delay cancer diagnoses [Zacharaki et al., 2009; Mangasarian et al., 1995], and because they have been shown to sometimes cause racist, sexist, and other discriminatory behaviors [Weber, 2012; Angwin et al., 2016]. Similarly, reinforcement learning algorithms have been proposed for applications where undesirable behavior can cause financial losses [Li et al., 2010], environmental damage [Houtman et al., 2013], and even death [Moore et al., 2010].

The regression algorithm that we designed attempts to minimize the mean squared error of its predictions while ensuring that, with high probability, a statistic of interest, $g(\theta)$, of the returned solution, $\theta = a(D)$, is bounded. The definition of this statistic can be chosen by the user to capture his or her desired definition of undesirable behavior (e.g., the expected financial loss that results from using a given solution, $\theta$). Importantly, the user of the algorithm may not know the value of this statistic for any solution. We must therefore provide the user with a way to tell our algorithm the statistic that he or she would like to bound, without requiring the user to provide the value of the statistic for any solution (see the second step of our framework). To achieve this, we allow the user to specify a sample statistic, and we define $g$ so as to ensure that the expected value of the sample statistic is bounded with high probability.

The creation of a regression algorithm (the third step of the framework) with the properties specified during the first two steps of the framework is challenging. This is to be expected given the shifted burden discussed previously. In the supplemental document we provide a detailed description of how we performed the third step of our framework for both our regression and reinforcement learning algorithms. At a high level, these algorithms were created by using statistical tests like Student's $t$-test and Hoeffding's inequality to transform sample statistics into bounds on the probability that $g(a(D)) > 0$, i.e., into bounds on the probability of undesirable behavior. Importantly, these two algorithms are not the contribution of this work—they are merely examples to prove that the design of algorithms using our framework is possible and tractable.

We applied a variant of our Seldonian regression algorithm to the problem of predicting how well an applicant to a university would perform if admitted, using a user-selected sample statistic that captures one form of discrimination.

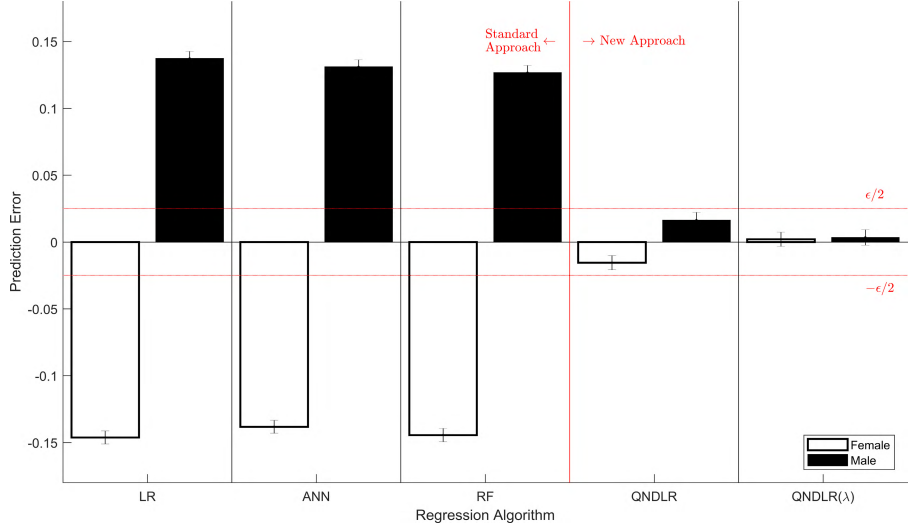

Figure 1: We used five different regression algorithms to predict students' *grade point averages* (GPAs) during their first three semesters at university based on their scores on nine entrance exams. We used real data from 43,303 students. Here, the user-selected definition of undesirable behavior corresponds to large differences in mean prediction errors (predicted GPA minus observed GPA) for applicants of different genders. This plot shows the mean prediction errors for male and female students when using each regression algorithm, and includes standard error bars. We used three standard algorithms: least squares linear regression (LR), an artificial neural network (ANN), and a random forest (RF), and two variants of our Seldonian algorithm: QNDLR and QNDLR($\lambda$). All of the standard methods tend to drastically over-predict the performance of male students and under-predict the performance of female students, while the two variants of our Seldonian regression algorithm do not. In particular, our algorithms ensure that, with approximately 95% probability, the expected prediction errors for men and women will be within $\epsilon = 0.05$, and both effectively preclude the sexist behavior that was exhibited by the standard algorithms.

Figure 1 presents the results of this experiment, which show that commonly used regression algorithms designed using the standard approach discriminate against female students, and that the user can easily limit this sexist behavior using our Seldonian regression algorithm.

Next we used our framework to design a Seldonian reinforcement learning algorithm [Sutton and Barto, 1998]: one that, unlike regression and classification algorithms, makes a sequence of dependent decisions. In this context, a solution, $\theta$, is called a *policy*; a *history*, $H$, (a random variable) denotes the outcome of using a policy to make a sequence of decisions; and the available data, $D$, is a set of outcomes produced by some initial policy, $\theta_0$. Since it is Seldonian, our algorithm searches for an optimal policy while ensuring that $\Pr(g(a(D)) \leq 0) \geq 1 - \delta$. The algorithm we designed is compatible with $g$ of the form: $g(\theta) \coloneqq \mathbf{E}[r'(H)|\theta_0] - \mathbf{E}[r'(H)|\theta]$, where the user selects $-r'(H)$ to measure his or her own definition of how undesirable the outcome $H$ is. That is, with probability at least $1 - \delta$, the algorithm will not output a policy, $\theta$, that increases the user-specified measure of undesirable behavior. Notice that the user need only be able to recognize undesirable behavior to define $r'$; the user does not need to know the distributions over trajectories, $H$, that result from applying different policies. For example, the user might define $r'(H) = -1$ if undesirable behavior occurred in $H$, and $r'(H) = 0$ otherwise.

We applied our Seldonian reinforcement learning algorithm to the problem of automatically improving the treatment policy (treatment regimen) for a person with type 1 diabetes [Bastani, 2014]. In this application, a policy, $\theta$, determines the amount of insulin that a person should inject prior to eating a meal, and each outcome, $H$, corresponds to one day. We defined $-r'(H)$ to be a measure of the prevalence of *hypoglycemia* (dangerously low blood sugar levels) in the outcome $H$. Figure 2 shows the result of applying both our Seldonian algorithm and a similar algorithm designed using the standard approach. For this experiment we used a detailed metabolic simulator [Dalla Man et al., 2014].

Given the recent surge of real-world machine learning applications and the corresponding surge of potential harm that they could cause, it is imperative that machine learning algorithms provide their users with an effective means for controlling behavior. To this end, we have proposed a new framework for designing machine learning algorithms and shown how it can be used to construct algorithms that provide their users with the ability to easily (that is, without requiring additional data analysis) place limits on the probability that the algorithm will produce any specified undesirable behaviors. Using this framework, we designed two algorithms and applied them to important high-risk real-world problems, where they precluded the sexist and harmful behavior exhibited when using algorithms designed using the standard approach. Algorithms designed using our framework are not just a replacement for machine learning algorithms in existing applications—it is our hope that they will pave the way for new applications for which the use of machine learning was previously deemed to be too risky.

In the remainder of this document we provide additional details and support for the claims that we have made. In §3 we discuss the standard approach for designing machine learning algorithms in more detail. In §4 we discuss the limitations of the standard approach that we address with our new framework. In §5 we provide a detailed description of Seldonian optimization problems—the problem formulation at the foundation of our framework. In §6 we give examples of how Seldonian regression algorithms can be designed and applied to limiting sexist behavior when applied to real-world data. In §7 we give an example of how a Seldonian reinforcement learning algorithm can be designed and applied to automatically personalizing treatment policies for type 1 diabetes.

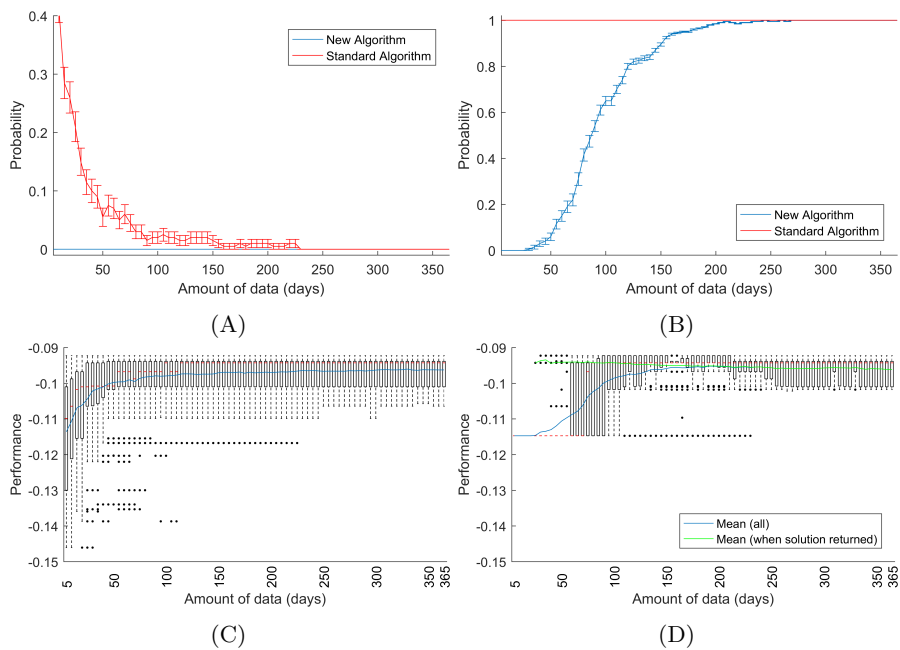

Figure 2: Results from diabetes treatment simulations, averaged over 200 trials. (**A**) Probability that each method returns treatment policies that increase the prevalence of hypoglycemia. The algorithm designed using the standard approach often proposed policies that increased the prevalence of hypoglycemia, and thus could triple the five year mortality rate [McCoy et al., 2012], even though it used an objective function (reward function) that penalized instances of hypoglycemia. By contrast, across all trials, our Seldonian algorithm was *safe*—it never changed the treatment policy in a way that increased the prevalence of hypoglycemia. (**B**) Probability that each method returns a treatment policy that differs from the initial treatment policy. Our Seldonian algorithm was able to safely improve upon the initial policy with just 3–6 months of data—a very reasonable amount [Bastani, 2014]. (**C**) The distribution of the expected returns (objective function values) of the treatment policies returned by the standard algorithm. The blue line depicts the sample mean, and the dashed red lines are the medians. All points below $-0.115$ correspond to cases where the standard algorithm decreased performance and produced undesirable behavior (an increase in the prevalence of hypoglycemia). (**D**) Similar to (C), but showing results for our new algorithm. The green line is the average of the performance when the algorithm produced a treatment policy that differs from the initial treatment policy. This line is not in (C) because the standard algorithm always produced policies that differed from the initial policy. Notice that all points are above $-0.115$, indicating that our algorithm never produced undesirable behavior.

## 2 Notation

In this document all random variables are denoted using uppercase letters, and expectations, $\mathbf{E}[\cdot]$, are always with respect to all of the random variables within the brackets unless otherwise specified. We use uppercase calligraphic letters to denote sets (except that $\Theta$ will also denote a set). We use lowercase letters to denote indices, elements of sets, and functions. We use $\mathbb{N}_{\geq 0}$ and $\mathbb{N}_{>0}$ to denote the natural numbers including zero and not including zero, respectively. We use $M^\intercal$ to denote the transpose of a matrix $M$, and we use column vectors rather than row vectors. We also reserve the symbol $n$ to always denote the number of constraints in an optimization problem (Seldonian or otherwise) and $m$ to denote the number of available data points.

## 3 The Standard Approach for Designing Machine Learning Algorithms

When designing a machine learning algorithm using the current standard approach, the first step is to define, mathematically, what the algorithm should try to do—the goal of the algorithm. At a sufficiently abstract level, this goal can be expressed the same way for almost all machine learning problems: find a *solution*, $\theta^\star$, within some *feasible set*, $\Theta$, that maximizes an *objective function*, $f : \Theta \to \mathbb{R}$. That is, the algorithm should search for an optimal solution,

$$\theta^\star \in \arg\max_{\theta \in \Theta} f(\theta).$$

To ground this abstract description of how the goal of an algorithm can be specified, consider the problem of designing a simple one-dimensional regression algorithm. We can begin by specifying the feasible set, $\Theta$, to be a set of functions. Each function, $\theta \in \Theta$, takes a real number as input and produces as output a real number, that is, $\theta : \mathbb{R} \to \mathbb{R}$. Let $X$ and $Y$ be correlated random variables, and our goal be to estimate $Y$ given $X$. We might then define the objective function to be the negative *mean squared error* (MSE):

$$f(\theta) \coloneqq -\mathbf{E}\left[(\theta(X) - Y)^2\right].$$

This completes the formal specification of *what* our algorithm should do, and so we can begin working on *how* our algorithm should do it. For example, we might have our algorithm construct an estimate, $\hat{f}$, of $f$ using data—$m$ realizations of $(X, Y)$, that is, $(x_i, y_i)$ for $i = 1, \ldots, m$. For example, $\hat{f}$ could be the negative *sample mean squared error*:

$$\hat{f}(\theta) \coloneqq -\frac{1}{m} \sum_{i=1}^{m} (\theta(x_i) - y_i)^2,$$

and our algorithm could return an element of $\arg\max_{\theta \in \Theta} \hat{f}(\theta)$.

Similarly, when designing a *reinforcement learning algorithm*, we might define $\Theta$ to be a set of policies for a Markov decision process and $f(\theta)$ to be an objective function like the *expected discounted return* [Sutton and Barto, 1998] or *average reward* [Mahadevan, 1996] of the policy $\theta$. In this case, we might create an algorithm that maximizes an estimate, $\hat f$, of $f$, constructed from data (sample state-transitions and rewards), or we might create an algorithm that, like *Q-learning* [Watkins, 1989], indirectly maximizes $f$ by optimizing a related function. When designing a *classification algorithm*, $\Theta$ would be a set of classifiers and $f$ a notion of how often the classifier selects the correct labels [Boser et al., 1992; Breiman, 2001; Krizhevsky et al., 2012]. When designing an *unsupervised learning algorithm*, we might define $\Theta$ to be a set of statistical models and $f(\theta)$ to be a notion of how similar $\theta$ is to a target statistical model [Dempster et al., 1977].

# 4 Limitations of the Standard Approach

Once a machine learning *researcher* has designed a machine learning algorithm, it can be used during the creation of an *agent*—a machine that uses implementations of one or more machine learning algorithms to solve a particular problem. The *user* of the algorithm—the person creating the agent—can be nearly anyone, from a child using LEGO Mindstorms, to a businessperson using Microsoft Excel to fit a line to data points, to a (non-computer) scientist using data analysis tools to analyze research data, to a machine learning researcher using a reinforcement learning algorithm for part of the controller of an autonomous vehicle. The primary limitation of the standard approach for designing machine learning algorithms is that it does not make it easy for the user of a machine learning algorithm (who may or may not be a machine learning expert) to specify and regulate the desirable and undesirable behaviors of the agent.

The user of the machine learning algorithm usually has the freedom to make a few decisions when applying a machine learning algorithm, and these decisions can impact the solution that the algorithm returns. It is through these decisions that the user can constrain the behavior of an agent using a machine learning algorithm. Three of the most common decisions that the user can make are what the feasible set, $\Theta$, and objective function, $f$, should be, and which machine learning algorithm to use. Although in principle there might be definitions of $\Theta$ and $f$ that cause the agent to never produce undesirable behaviors, in practice there is no way for the user to know these definitions without performing additional data analysis that can be challenging even for a machine learning researcher.

As an example, consider a reinforcement learning application where an artificial neural network is used to control a robot. In this context, the feasible set, $\Theta$, is a set of neural networks (typically with the same structure, but different weights), each of which would cause the robot to behave differently. Reinforcement learning algorithms can be used to search for an artificial neural network within $\Theta$ that performs well in terms of some user-defined performance measure. The user of a reinforcement learning algorithm might want to implement Asimov's first law of robotics, which essentially states that a robot may not harm a human [Asimov, 1951]. However, the user of the algorithm typically does not know whether any particular artificial neural network, $\theta$, will cause the robot to harm a human or not. This means that the user of the algorithm cannot specify $\Theta$ to only include artificial neural networks that produce safe behavior. Additionally, since most reinforcement learning algorithms are not guaranteed to produce an optimal solution given finite data, simply adding a penalty into the objective function to punish harming a human does not preclude the algorithm from returning a suboptimal solution (artificial neural network) that causes harm to a human. This means that the user of the algorithm has no easy way to constrain the behavior of the agent—the user of the algorithm must have deep knowledge of the environment that the robot will be faced with, what each artificial neural network, $\theta \in \Theta$, does, and how the reinforcement learning algorithm works, in order to ensure that the reinforcement learning algorithm will not cause the robot (agent) to harm a human.

In §4.1 we present an illustrative example of a problem where it is difficult to ensure that an agent is well-behaved. In §4.2 through §4.8 we then discuss the limitations of current techniques for ensuring that an agent is well-behaved, using the aforementioned illustrative example to ground our discussion. The example that we present shows how undesirable behavior can occur even when using one of the most widely used and well-studied data analysis tools: linear regression.

## 4.1 An Illustrative Example

To ground our subsequent discussions, we propose a simple linear regression problem. For this problem, our goal is to predict the aptitudes of job applicants based on numerical values that describe the qualities of their résumés. Although later we consider a similar problem using advanced regression algorithms and real data, here we use easily reproduced synthetic data and only consider linear regression algorithms. We will show how linear regression algorithms designed using the standard approach can result in predictions of applicant performance that systematically discriminate against a group of people (such as people of one gender or race).

Let each applicant either be in a set, $\mathcal{A}$, or another set, $\mathcal{B}$. For example, $\mathcal{A}$ could be the set of all possible female applicants and $\mathcal{B}$ could be the set of all possible male applicants. We refer to the group that the applicant belongs to as his or her *type*. We are given a *training set* that contains data describing the résumés of $m = 1{,}000$ previous applicants, the applicants' actual aptitudes, and their types. For each $i \in \{1, \ldots, m\}$, let $x_i \in \mathbb{R}$ be a number describing the quality of the $i^{\text{th}}$ applicant's résumé, $y_i \in \mathbb{R}$ be a measure of the applicant's actual aptitude (which we would like to predict given $x_i$), and let $t_i \in \{0, 1\}$ be an indicator of what type the applicant is—$t_i = 0$ if the applicant is in $\mathcal{A}$ and $t_i = 1$ if the applicant is in $\mathcal{B}$. For simplicity, we assume that all terms, $x_i$ and $y_i$, have been normalized so that they each tend to be inside the $[-3, 3]$ interval.

We are tasked with using the training set to find a *linear function*, $\hat{y}(x,\theta) \coloneqq \theta_1 x + \theta_2$, that predicts $y$ given $x$, where $x$ is the number describing the quality of a new applicant's résumé, $y$ is the unknown aptitude of this new applicant, and the vector $\theta = [\theta_1, \theta_2]^\intercal$ is a *weight vector* in the feasible set, $\Theta = \mathbb{R}^2$. Although each applicant's type is available in the training set, we do not assume that we will know a *new* applicant's type: résumés do not typically include applicants' genders or ethnicities. If our predictor is used to filter actual résumés submitted to a company so that only a subset of applications receive human review—an existing application of machine learning [Weber, 2012; Miller, 2015]—then to comply with anti-discrimination laws, we might want to ensure that our estimator does not produce racist or sexist behavior—that it does not discriminate against people in $\mathcal{A}$ or $\mathcal{B}$. There are many ways that we might choose to define undesirable discriminatory behavior—here we choose one. Our goal is not to propose that this single definition of discriminatory behavior captures all possible types of discrimination, but to present an example of behavior that cannot easily be precluded when using algorithms designed using the standard approach for designing machine learning algorithms.

Intuitively, we define *undesirable discriminatory behavior* to be when a line produces predictions that are, on average, too high for people of one type and too low for people of the other type. Notice that we do not necessarily consider it to be undesirable discriminatory behavior if a line produces larger predictions, on average, for people of one type, since people of one type might actually have higher aptitudes on average. However, if the line found by our algorithm over-predicts the performance of men by 10% and under-predicts the performance of women by 10%, on average, then we would say that the line discriminates against women. To formalize this notion, we define the *discriminatory statistic*, $d(\theta)$ to be:

$$d(\theta) \coloneqq \underbrace{\mathbf{E}\Big[u\left(\hat{y}(X,\theta) - Y\right) \Big| T=0\Big]}_{(a)} - \underbrace{\mathbf{E}\Big[u\left(\hat{y}(X,\theta) - Y\right) \Big| T=1\Big]}_{(b)}, \qquad (3)$$

where $X, Y$, and $T$ are random variables that denote the numerical measure of résumé quality, actual aptitude, and type of an applicant, $u : \mathbb{R} \to \mathbb{R}$ is a *utility function*, and where (a) and (b) indicate how much the estimator over-predicts on average for people in $\mathcal{A}$ and $\mathcal{B}$, respectively. Here the utility function, $u$, can be used to determine the relative importance of different amounts of over and under-prediction. For example, the impact of an *over*-prediction by $\Delta$ may not have the same magnitude as the impact of an *under*-prediction by $\Delta$, and so one may define $u$ such that $u(\Delta) \neq -u(-\Delta)$ for some $\Delta \in \mathbb{R}$. Furthermore, the difference between an error of $\Delta$ and $\Delta + \epsilon$ (for some small $\Delta$ and $\epsilon$) may be more significant than the difference between an error of $100\Delta$ and $100\Delta + \epsilon$, and so $u$ may not be a linear function: $|u(\Delta) - u(\Delta+\epsilon)| \gg |u(100\Delta) - u(100\Delta+\epsilon)|$. Although nonlinear utility functions can mimic the way that humans gauge utility [Grayson, 1960], for simplicity here, we assume that $u(\Delta) = \Delta$—the utility function is the identity function.

Given a training set that contains data from 1,000 past applicants, we would like to apply a linear regression algorithm to get accurate predictions of applicant aptitudes while simultaneously ensuring that the absolute value of the discriminatory statistic, which we call the *absolute discriminatory statistic*, $|d(\theta)|$, is small. An example of what the training set might look like, along with the line produced by a popular linear regression algorithm—least squares linear regression—is provided in Figure 3.

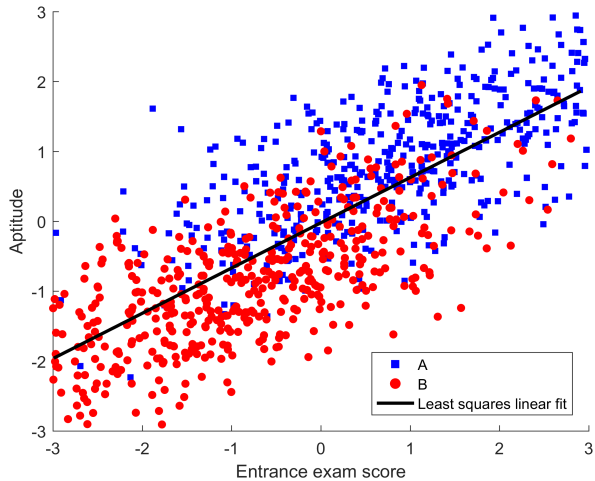

Figure 3: An example of a training set from the illustrative example. This figure depicts all of the information that the user is given when applying a linear regression algorithm. That is, the user does *not* know the true underlying distribution of the data, but rather only has these $m = 1{,}000$ samples. The black line is the line produced by least squares linear regression.

To further ground this example, we explicitly define additional information about the problem that would *not* be known to the user of a linear regression algorithm: the true underlying distributions of $X$, $Y$, and $T$. Specifically, let future applicants be in $\mathcal{A}$ and $\mathcal{B}$ with equal probability and let the training set have an equal number of applicants of each type. Let $Y \sim \mathcal{N}(1, 1)$ if $T = 0$ and $Y \sim \mathcal{N}(-1, 1)$ if $T = 1$, where $\mathcal{N}(\mu, \sigma^2)$ denotes the normal distribution with mean $\mu$ and standard deviation $\sigma$. Let $X \sim \mathcal{N}(Y, 1)$—that is, an applicant's résumé quality is equal to their true aptitude, plus random noise with a standard normal distribution. The training set contains $m = 1{,}000$ realizations of $X, Y$, and $T$. That is, $D = \{(X_i, Y_i, T_i)\}_{i=1}^{m}$.

## 4.2 Potential Remedy: Rely on the Impartiality of the Algorithm

It may be tempting to ignore the problem of undesirable behavior entirely. Computers do not have an inherent desire to produce undesirable behavior—for example, to harm humans or to be racist or sexist—and the user of an algorithm typically will not provide the algorithm with direct incentives to produce undesirable behavior. However, a lack of direct incentives to produce undesirable behavior does not preclude undesirable behavior. Consider two solutions, $\theta_1 \in \Theta$ and $\theta_2 \in \Theta$, where $\theta_1$ produces undesirable behavior that is unrelated to the objective specified by the objective function, while $\theta_2$ does not produce the undesirable behavior. If the objective function takes the same values for both solutions, and their values are optimal, then the machine learning algorithm might return either. Alternatively, there might be an unanticipated correlation between the objective function and the undesirable behavior that causes $f(\theta_1) > f(\theta_2)$, in which case the machine learning algorithm might often return $\theta_1$. This shows that we cannot rely on the impartiality of the machine learning algorithm—the fact that it was not instructed to produce undesirable behavior.

Consider our illustrative example. Here we might choose to select the line that minimizes the sample mean squared error—the least squares fit. Since the algorithm's goal is to make accurate predictions, we might expect it to be impartial to whether people are in $\mathcal{A}$ or $\mathcal{B}$, and so it should not tend to produce discriminatory behavior—its impartiality should make it fair to all people. However, this is not the case. This is an example where there is an unanticipated correlation between the objective function and the undesirable behavior (large values of the discriminatory statistic). We generated 10,000 independent training sets (each containing data from $m = 1{,}000$ applicants), computed the least squares linear fits for each data set, and computed the true discriminatory statistic for each of the resulting lines (using our knowledge of the true distribution of the data). The mean discriminatory statistic was $-0.67$—a large amount of discrimination in favor of people in group $\mathcal{B}$, given that applicant aptitudes tend to be in the range $[-3, 3]$. Figure 4 shows the least squares linear fits for all of the 10,000 trials.

## 4.3 Potential Remedy: Determine and Combat the Root Causes of Undesirable Behavior

When undesirable behavior occurs, it is natural to wonder what the root cause of the behavior was. Was the undesirable behavior caused by improper use of a machine learning algorithm? Would a different way of using an algorithm designed using the standard approach preclude the undesirable behavior? In general: what could have been done differently by the user of the algorithm to cause the agent to not produce the undesirable behavior? This is exactly the question that we do not want the user of an algorithm to have to answer, since it requires detailed knowledge of the problem to answer, and can easily be answered incorrectly. At a high level, our argument is that this process places an undue burden on the users of machine learning algorithms (who may not be machine learning experts), and this burden should be shifted away from the user.

The severity of this burden is obvious for complicated problems: the user of a complicated reinforcement learning algorithm that tunes the weights in a neural network controlling a robot cannot be expected to know which weights or settings of the algorithm would result in the robot eventually harming a human. Perhaps less obviously, this problem impacts even simple data analysis algorithms. To show this, consider our illustrative example, and what would be required of the (possibly non-technical) user of a linear regression algorithm who wishes to understand and mitigate the root causes of undesirable discriminatory behavior.

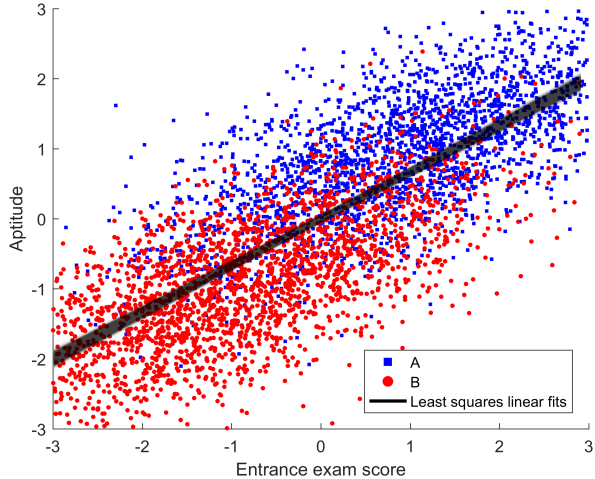

Figure 4: Linear fits using least squares for 10,000 trials using transparent lines to show the distribution of lines. The background contains 5,000 realizations of $(X, Y, T)$ to give an idea about the true underlying distribution of the data. Notice that the lines tend to over-predict for red points (people of type $T = 1$) and under-predict for blue points (people of type $T = 0$).

One possible cause of discriminatory behavior might be an imbalance in the training data. If more applicants from group $\mathcal{B}$ are observed than applicants from $\mathcal{A}$, then we might expect the regression algorithm to favor lines that produce more accurate predictions for people in $\mathcal{B}$, even if it results in worse predictions for people in $\mathcal{A}$. However, this is not the cause of discriminatory behavior in this example. Note that the discriminatory statistic, $d(\theta)$, is not necessarily related to how accurate an estimator is—it could still be nondiscriminatory $(d(\theta) \approx 0)$ as long as it does not consistently over or under-predict for applicants of each type. In our example we defined the underlying applicant distribution so that there is no minority group. This shows that imbalance in the training data is not the underlying cause of discrimination in our example, and therefore that guaranteeing that a data set is balanced is not sufficient to preclude discrimination.

Another possible cause of discriminatory behavior might be bias in the data set. If the training data was biased so that it over-reported the aptitudes of applicants in $\mathcal{A}$ relative to their true aptitude (which is used when computing the discriminatory statistic), then the regression algorithm would also be biased towards discriminating in favor of applicants in $\mathcal{A}$. However, in this synthetic example, the training and testing data come from the same distribution, and so there is no additional bias in the training data.

Another possible cause of discriminatory behavior might be our use of a linear estimator. Perhaps the minimum mean squared error estimator is not linear, and the closest linear approximation happens to discriminate. This suggests that using a more representative class of estimators might mitigate undesirable discriminatory behavior. Since we know the true underlying distribution of applicants, this is straightforward to test. In Appendix A we show that, for any résumé quality, $x$, the estimate of aptitude that minimizes the mean squared error (not just the *sample* mean squared error) is $\hat{y}(x) = \frac{2}{3}x$. That is, the minimum mean squared error estimates for each possible résumé quality, $x$, is given by a linear function, and so the optimal estimator (in terms of mean squared error) is in our function class.

Another possible cause of discriminatory behavior might be the use of finite amounts of data. Since the training data is finite and generated randomly, different data sets will result in different linear fits. Perhaps the linear fits are in some way centered around an estimator that does not discriminate, but the way that they vary around this estimator causes the absolute value of the mean discriminatory statistic to be large. Since we know the underlying distribution of applicants, we can check this by solving for the single estimator that would be produced if given an infinite amount of data. As discussed previously, this estimator is almost surely $\hat{y}(x) = \frac{2}{3}x$. Using our knowledge of the underlying distribution of applicants again, we can compute the discriminatory statistic for this line, $d([\frac{2}{3}, 0]^\intercal) = -0.67$. This means that, even given an infinite amount of data so that the linear fits have no variance, the linear regression algorithm still discriminates.

Another possible cause of discriminatory behavior could stem from the decision of whether the regression algorithm's predictions can depend on the applicant's type (whether they are in $\mathcal{A}$ or $\mathcal{B}$). First, one might expect that if the algorithm is not aware of the applicants' types, it could not possibly discriminate. However this is not the case: in our illustrative example the linear regression algorithm is blind to the type of each applicant, but still discriminates. Second, one might expect the opposite: that if applicants' performances depend on their types, then the algorithm should be aware of the applicants' types so that it can make fair predictions. For example, we could apply the least squares regression algorithm twice—once to the data from people in $\mathcal{A}$, and once to the data from people in $\mathcal{B}$. To predict the aptitude of a new applicant, we apply the first line if the applicant is in $\mathcal{A}$ and the second line otherwise.

Notice that this approach does not allow for generalization across types, which could drastically reduce data efficiency, particularly when there are multiple types. However, despite this drawback, this approach can be effective and will preclude discrimination provided that three conditions are met. **First**, the regression algorithm must have access to new applicants' types, which may not be the case—the résumés of new applicants likely will not include their genders or ethnicities.

**Second**, sufficient data must be available. If insufficient data is available, then this approach might produce discriminatory behavior because the training data could be an unlikely sample that does not reflect the real distributions of $X, Y$, and $T$. Determining the necessary amount of data for this scheme to limit discrimination with high probability would require the user of the algorithm to perform additional data analysis. Furthermore, when there is insufficient data, one might be tempted to return to the argument that this scheme results in no inherent bias in the predictions, and so we can rely on its impartiality. That is, even with small amounts of data, perhaps at least the expected (not-absolute) discriminatory statistic will be zero. However, this is not always the case—Figure 5 provides an example of a different distribution of applicants where using small amounts of data can still cause the expected discriminatory statistic to be large.
**Third**, the utility function must be the identity function. For every linear regression problem (provided that one feature is a non-zero constant—one weight encodes the $y$-intercept) the line that minimizes the (true, not sample) mean squared error causes the mean error to also be zero, and so this scheme results in a discriminatory statistic close to zero given enough data. However, the line that minimizes the mean squared error does not necessarily cause the mean *utility* of the error to also be zero, and so utility functions other than the identity function can result in this scheme producing discriminatory lines even given infinite data.

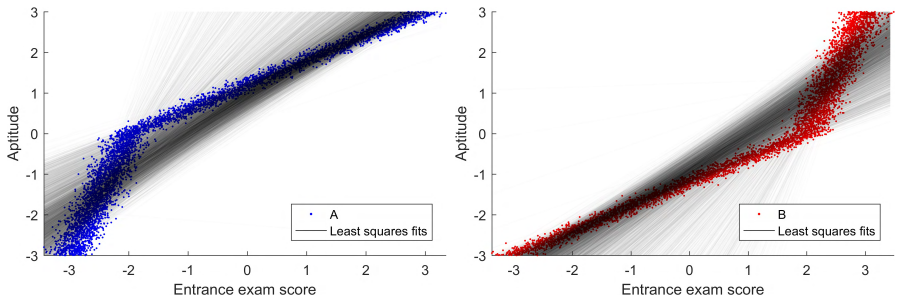

Figure 5: Different distributions for people of types $\mathcal{A}$ (left) and $\mathcal{B}$ (right), such that using independent linear regressors for people of each type, and training data from five people of each type, results in an average discriminatory statistic (over 1,000 trials) of 0.42.

It should now be clear that the root cause of the discriminatory behavior is not obvious, and could easily be overlooked or incorrectly attributed (particularly if the user of the algorithm is not trained in data analysis methods). In our example, the real root cause of discriminatory behavior when using ordinary least squares linear regression comes from the fact that the objective function calls for minimizing mean squared error, which is at odds with minimizing the discriminatory statistic. To minimize the discriminatory statistic (ideally, to make it zero) requires the machine learning algorithm to return a line that has higher mean squared error than other lines. Intuitively, the function that minimizes the (sample) mean squared error can still have mean errors that are different for people of different types. In this example the mean error of the least squares fit tends to be positive for people with low aptitudes and negative for people with high aptitudes. People in $\mathcal{B}$ tend to have lower aptitudes, and so minimizing the mean squared error results in discrimination in favor of people in $\mathcal{B}$.

In summary, although it might be possible to determine and combat the root causes of undesirable behavior, doing so can be difficult, error prone, and could require data analysis, even for simple well-understood algorithms like linear regression. However, so far we have primarily discussed algorithms that were not designed with the explicit intent to allow the user to control behavior. Some algorithms were designed using variants of the standard approach that were created specifically to provide guarantees about the behavior of agents. In the following subsections we review some of these methods.

## 4.4 Potential Remedy: Add Hard Constraints

As we mentioned earlier, some machine learning algorithms allow the user to place constraints on $\Theta$, such as the simplex algorithm for linear programs, which allows (actually, requires) the user to define the feasible set using linear constraints [Dantzig et al., 1955]. Constraints on the feasible set have been proposed before as a means for controlling agent behavior. For example, Thomas et al. [2013] propose a reinforcement learning algorithm that allows the user to specify a "safe" set of policies, and guarantees that the algorithm will only ever deploy policies from within this safe set. However, Thomas et al. [2013] sidestep the question of how this safe set of policies could be computed by assuming that it is provided *a priori*. Herein lies the problem with these methods: determining which solutions are "safe"—which solutions do not result in undesirable behaviors—can be difficult, requires detailed knowledge of the problem at hand, and can sometimes be impossible. Although some work has considered how hard constraints can easily be specified by a user, e.g., by providing examples of desirable and undesirable behavior [Irani, 2015], these approaches do not provide practical guarantees about the quality of the constraints that they produce.

Consider our illustrative example. Without constraints, the feasible set is the set of all possible pairs of weights that define a line: $\Theta = \mathbb{R}^2$. We might wish to define the set of safe solutions, $\mathcal{S} \subseteq \mathbb{R}$, to be all of the solutions that result in discriminatory statistics with magnitude at most some small value, $\epsilon$. That is, $\mathcal{S} \coloneqq \{\theta \in \mathbb{R}^2 : |d(\theta)| \leq \epsilon\}$. Our least squares algorithm could then be constrained to only consider solutions in $\mathcal{S}$. The problem here is that we cannot know $\mathcal{S}$ without knowledge of the underlying distribution of applicants: their types, aptitudes, and résumé qualities.

However, constraints on the feasible set can be an effective means for precluding some specific definitions of undesirable behavior, provided that the user has access to detailed knowledge of the problem. For example, control theoretic algorithms have been developed that ensure (sometimes with high probability, rather than surely) that a system will never enter a predefined *unsafe* set of states [Tomlin, 1998; Oishi et al., 2001; Perkins and Barto, 2003; Mitchell et al., 2005;

Hans et al., 2008; Arvelo and Martins, 2012; Akametalu et al., 2014; Zilberstein, 2015]. These methods can be viewed as using a feasible set that only includes controllers that avoid unsafe regions of state space.

## 4.5 Potential Remedy: Add Soft Constraints

Rather than insert constraints into $\Theta$, we might consider modifying $f$ (or an estimate, $\hat{f}$, thereof) so that it penalizes undesirable behavior. These sorts of penalties in the objective function are sometimes called *soft constraints* [Boyd and Vandenberghe, 2004], because the algorithm is given incentive to satisfy them (to preclude the undesirable behaviors) but has the freedom to violate the constraints if it results in a large improvement of the original objective function. Soft constraints are also sometimes called *penalty functions* and are related to *barrier functions* [Nocedal and Wright, 2006], which penalize solutions that are close to the boundary of a feasible set.

Although soft constraints can sometimes be effective, they have a significant drawback: they require a parameter, $\lambda \in \mathbb{R}_{\geq 0}$ that scales the importance of the soft constraint relative to the primary objective. If this parameter is set too far to one extreme, then the algorithm will ignore the soft constraint and focus entirely on the primary objective, which means that undesirable behavior could manifest. If this parameter is set too far to the other extreme, then the algorithm will ignore the primary objective and focus solely on ensuring that undesirable behavior does not occur. Properly selecting $\lambda$ is not simple, and requires additional data analysis that may be unreasonable to expect from a user not trained in data analysis.

Consider, for instance, our illustrative example. Here we might introduce a soft constraint so that the objective calls for the simultaneous minimization of the mean squared error and the absolute value of the discriminatory statistic. That is, if $\Theta = \mathbb{R}^2$ so that each $\theta \in \Theta$ is a possible weight vector defining a line, then:

$$\begin{aligned} f(\theta) &\coloneqq -\operatorname{MSE}(\theta) - \lambda |d(\theta)| \\ &= -\mathbf{E}\left[(\hat{y}(X, \theta) - Y)^2\right] - \lambda \bigg| \mathbf{E}\Big[\hat{y}(X, \theta) - Y \Big| T = 0\Big] \\ &\quad - \mathbf{E}\Big[\hat{y}(X, \theta) - Y \Big| T = 1\Big]\bigg|. \end{aligned}$$

We might then define our data-based estimate of $f(\theta)$ to use the sample mean squared error and sample discriminatory statistic:

$$\hat{f}(\theta) := -\frac{1}{m}\sum_{i=1}^{m}(\hat{y}(x_i,\theta) - y_i)^2 - \lambda \left| \left( \frac{1}{\sum_{i=1}^{m}(1-t_i)} \sum_{i=1}^{m} (1-t_i)\,(\hat{y}(x_i,\theta) - y_i) \right) \right.$$

$$\left. - \left( \frac{1}{\sum_{i=1}^{m} t_i} \sum_{i=1}^{m} t_i\,(\hat{y}(x_i,\theta) - y_i) \right) \right|. \tag{4}$$

Although in some cases it is reasonable to expect the user of a machine learning algorithm to be able to specify a soft constraint of this form, it is typically not reasonable to expect the user to be able to select the value of $\lambda$ properly. To show this, consider Figure 6, which depicts the mean discriminatory statistic and MSE of solutions, $\hat{\theta}^\star \in \arg\max_{\theta \in \Theta} \hat{f}(\theta)$, when using (4) with various $\lambda$. As $\lambda$ increases, MSE increases as well, since the objective function places increasing weight on avoiding discrimination at the cost of MSE. The left plot of Figure 6 depicts the discriminatory statistic of the solutions produced using various $\lambda$. As $\lambda$ increases, the discriminatory statistic decreases, as expected.

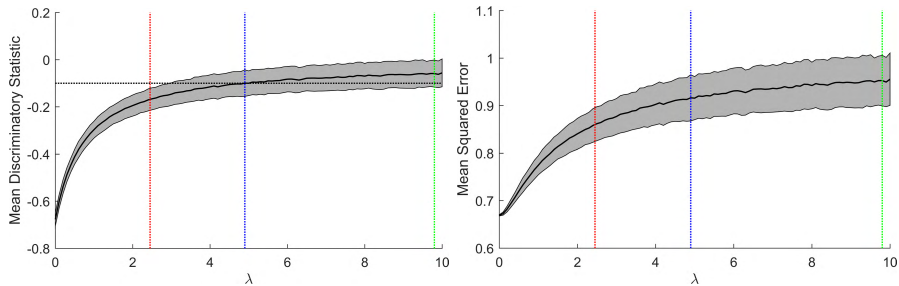

Figure 6: Mean discriminatory statistic (left) and mean squared error (right) produced by the solutions found using various $\lambda$ to specify the importance of a soft constraint. The dotted blue line is the smallest value of $\lambda$ for which the mean discriminatory statistic is less than 0.1 (that is, $\lambda = 4.0$), and the red and green line are half and double this value. Both plots are averaged over 1,000 trials (each from a new sampling of the training set) and the upper and lower standard deviation intervals are shaded (we report standard deviation because standard error bars are too small to be clearly visible).

In order to ensure that the expected discriminatory statistic is at most some value, $\epsilon$, we can use the plot on the left to find the smallest value of $\lambda$ that produces discriminatory statistics that are less than $\epsilon$ on average—$\lambda \approx 4.9$ if $\epsilon = 0.1$. However, in practice these plots are *not* available to the user—they were generated by additional data analysis (and also using large amounts of data that would not be available in practice). Consider what would happen if the user's choice of $\lambda$ is off by a factor of two. If the user selects $\lambda = 2.45$, then the absolute discriminatory statistic will be significantly larger than 0.1 on average, and so the user's desire to control behavior will be undermined. If the user selects $\lambda = 9.8$, then the absolute discriminatory statistic will be sufficiently small on average, but the resulting mean squared error will be unnecessarily large.

Ideally, the machine learning algorithm should use the available data to automatically optimize the value of $\lambda$ in a way that, when combined with finite sample bounds, provides the (possibly non-technical) user with confidence that the algorithm will not cause discriminatory behavior. Such an algorithm would be an instance of the type of algorithm that we propose—it shifts, from the user of the algorithm to the algorithm itself, the onus of understanding the problem at hand well enough to constrain behavior.

## 4.6 Potential Remedy: Multiobjective Methods

Soft constraints are one way of expressing the idea that the true objective is multifaceted: the agent should optimize some primary objective while also optimizing other objectives that measure the prevalence of undesirable behavior. *Multiobjective* optimization algorithms are algorithms designed specifically to deal with multiple (typically conflicting) objectives of this sort.

Modern multiobjective optimization algorithms tend to be based on the concept of the *Pareto frontier*. Intuitively, a solution, $\theta$, is said to be on the Pareto frontier if there does not exist another solution, $\theta' \in \Theta$ that causes any of the objectives to be increased (assuming that larger values are better for all objectives) without decreasing at least one of the other objectives. More formally, if $f_1, \ldots, f_k$ are $k$ objective functions, then the Pareto frontier is the set

$$P \coloneqq \left\{ \theta \in \Theta : \forall \theta' \in \Theta, \left( \exists i, \ f_i(\theta') > f_i(\theta) \implies \exists j, \ f_j(\theta') < f_j(\theta) \right) \right\}.$$

All of the solutions on the Pareto frontier provide a balance between the different objectives, and one should probably return a solution on the Pareto frontier (since any other solution could be improved with respect to at least one objective function without hindering performance with respect to any of the other objective functions).

However, while algorithms exist to compute the Pareto frontier for a wide variety of multiobjective machine learning problems [Messac et al., 2003; Smits and Kotanchek, 2005; Pirotta et al., 2014], knowing the Pareto frontier does not provide a complete solution. One must still decide which solution from the Pareto frontier to use—the user must still explicitly select a trade-off between the different objective functions. Consider our illustrative example, where we might use two objective functions: the negative mean squared error of the estimator, $-\operatorname{MSE}(\theta)$, and the negative absolute value of the discriminatory statistic, $-|d(\theta)|$. Here each line that results from using a soft constraint with any value of $\lambda$ is an element of the Pareto frontier, and so the user must still effectively determine how to select $\lambda$.

Later we present two linear regression algorithms, NDLR and QNDLR, that were designed using our new framework. Although these algorithms do not have a $\lambda$ parameter, they do have a different parameter, $\epsilon$. Specifically, they guarantee that (with high probability) the absolute value of the discriminatory statistic will be no larger than $\epsilon$. The difference between requiring the user to select $\lambda$ and requiring the user to select $\epsilon$ is subtle but important: the scale (or unit) associated with $\epsilon$ is one that can be understood by the user without any problem-specific data analysis—$\epsilon$ is the maximum difference in mean prediction errors, and so its associated scale is the scale of errors. By contrast, the scale (or unit) associated with $\lambda$ (how different sizes of $\lambda$ map to different discriminatory statistics) depends on the mean squared error of the solution, which typically is not known to the user, and the estimation of which requires additional data analysis. Thus we contend that it is easier for the user to select $\epsilon$ than it is for the user to select $\lambda$.

Furthermore, regardless of how large $\lambda$ is, using a soft constraint can still result in solutions, $\theta$, that discriminate significantly. This is due to the random nature of data: small data sets may not accurately represent the true underlying distribution of data. As a result, a soft-constrained algorithm may select a solution that has a *sample* discriminatory statistic of zero on the available data, but which actually has a large discriminatory statistic (discriminates when presented with new data). This behavior is evident in our later experiments (c.f., Figure 7B), wherein using small amounts of data and large values of $\lambda$, soft-constrained methods produce solutions that discriminate significantly. By contrast, with high probability, the algorithms that we design using our new framework do not discriminate regardless of how much data they are presented with. This is because they incorporate mechanisms that automatically determine whether the random nature of data is causing them to draw false conclusions.

## 4.7 Potential Remedy: Use Chance-Constraints

For some (perhaps most) applications, there may not be any solutions that preclude undesirable behavior with certainty. For these applications, requiring undesirable behaviors to never occur is too strict, and so we must weaken our requirements. This is what the *chance-constrained program* problem formulation does—it allows for constraints on the probability that a solution will result in undesirable behavior. Chance-constrained programs were pioneered by Charnes and Cooper [1959], Miller and Wagner [1965], and Prékopa [1970], and can be expressed formally as:

$$\arg\min_{\theta \in \Theta} f(\theta)$$
$$\text{s.t. } \forall i \in \{1,\ldots,n\}, \ \Pr(g_i(\theta, W_i) \leq 0) \geq 1 - \delta_i,$$

where $f : \Theta \to \mathbb{R}$, each $g_i$ is a deterministic real-valued function, each $W_i$ is a random variable, and each $\delta_i \in (0,1)$. Crucially, not only are $\Theta$, $g_i$, and $\delta_i$ all known, but the distribution of each $W_i$ is known. Also, typically $f$ is a convex function and $\Theta$ is a convex set. Chance-constrained programs have been heavily studied within the machine learning and optimization communities [Dentcheva et al., 2000; Nemirovski, 2012], even with the expressed intent of better controlling agent behavior [Xu and Mannor, 2011].

As an example of when chance-constraints might seem to be appropriate, consider the search for a neural network, $\theta \in \Theta$, that could be used to control a robot tomorrow. Intuitively, one can think of $\mathcal{W}$ as the set of possible environments (or *worlds*) that the robot could be faced with tomorrow, and $W_i \in \mathcal{W}$ is a random variable that denotes the world that the robot will be faced with tomorrow. There may not exist a solution, $\theta \in \Theta$, that guarantees with certainty that it will never harm a human regardless of which world, $w \in \mathcal{W}$ is it faced with tomorrow. Chance constraints allow us to define a "safe" set of solutions, $\mathcal{S} \subseteq \Theta$, such that each $\theta \in \mathcal{S}$ ensures that with high probability it will not result in harm to a human tomorrow, given that the world tomorrow, $W_i$, is drawn from some known distribution. That is, $\mathcal{S} = \{\theta \in \Theta : \forall i \in \{1, \ldots, n\}, \Pr(g_i(\theta, W_i) \leq 0) \geq 1 - \delta_i\}$, where $g_i(\theta, w) > 0$ denotes that, if the true world tomorrow happens to be $w \in \mathcal{W}$, then the solution, $\theta$, will result in harm to a human. One can view chance constraints as a language for specifying the hard constraint that the feasible solutions should be restricted to $\mathcal{S}$ rather than $\Theta$.

Although chance-constrained programs can be useful for many real problems [Houck, 1979; Gurvich et al., 2010; Wang et al., 2012], like other approaches based on hard constraints on the feasible set, their applicability is limited to problems where the user has significant knowledge about the problem at hand. Specifically, chance-constrained programs are a viable problem formulation when the user knows the distribution of $W_i$. However, for many problems this is the not the case. In our illustrative example, the random variable, $W_i$, might denote a single point, $(X, Y, T)$, it might denote a set of points (such as the entire training set), or it might denote the joint distribution over $(X, Y, T)$ if we assume that the distribution of applicants is itself sampled from some meta-distribution. In all of these cases, the user does *not* have access to the distribution of $W_i$, and therefore cannot construct $\mathcal{S}$.

There are several variants of chance constrained programs that weaken the assumption that the distribution of $W_i$ is known. For example, *ambiguous chance constrained programs* allow the distribution, $P$, of $W_i$ to be unknown, as long as it is within some set of possible distributions, $\mathcal{P}$, that is known [Erdoğan and Iyengar, 2006]. Ambiguous chance constrained programs then require the chance constraints to hold for all $P \in \mathcal{P}$. *Scenario approximation* methods can apply when $P$, the distribution of $W_i$, is neither known nor within a known set, $\mathcal{P}$, but samples (realizations of $W_i$) can be generated from $P$ [De Farias and Van Roy, 2004; Calafiore and Campi, 2005; Nemirovski and Shapiro, 2006]. In general, *stochastic programming* methods allow for many variants of optimization problems wherein there is uncertainty about some of the parameters of the optimization problem [Birge and Louveaux, 2011]. However, these variants tend to include strong assumptions about the form of the problem: primarily that the objective function and feasible set are convex, and that the distributions of random variables are restricted to a specific class (e.g., Gaussians). To the best of our knowledge, none of the variants are able to handle our illustrative example: they do not allow for the specification that the objective is to minimize the mean squared error of the linear fit, subject to the constraint that with high probability the absolute discriminatory statistic, $|d(\theta)|$, of the returned solution, $\theta$, is bounded by some small constant, $\epsilon$.

Although chance-constrained programs deal with the "probability that a solution produces undesirable behavior," which sounds related to what we might want to achieve, there is a subtle but important distinction between chance-constrained programs and what we will propose in this work. The key distinction has to do with the differences between **(1)** the probability that a solution will cause undesirable behavior and **(2)** the probability that a machine learning algorithm will produce a solution that has high probability of causing undesirable behavior. Notice that **(1)** is a property that is inherent to a solution, and which is unrelated to any available data: either a solution produces undesirable behavior with low probability or it does not. Chance constraints allow for constraints on **(1)**—they allow for the specification of $\mathcal{S}$ as defined earlier. Just as we cannot expect the user to know that a solution (e.g., neural network) will never produce undesirable behavior (cause a robot to harm a human), we cannot expect the user to know that a solution (neural network) will cause the probability of undesirable behavior to be low (the probability of harm to a human to be low). The user may not (and may never) have access to the information necessary to determine with certainty whether a solution is in $\mathcal{S}$ or not. However, we might expect that a machine learning algorithm could use data to reason about its belief about whether a solution is in $\mathcal{S}$, and so a machine learning algorithm might be able to use data to ensure that **(2)** is small. This is what we will propose: the user will provide the machine learning algorithm with a definition of undesirable behavior (but not any information about how likely any solution is to cause undesirable behavior), and the machine learning algorithm will then use data to ensure that **(2)** is small.

## 4.8 Potential Remedy: Use Robust or Risk-Sensitive Methods

Determining which solutions will produce undesirable behavior is difficult because there is typically some uncertainty about the environment that an agent will interact with. *Robust optimization* algorithms are algorithms that handle uncertainty about the environment in a conservative way: they favor solutions that work reasonably well across all of the possible environments, rather than a solution that is expected to be optimal for one particular environment [Ben-Tal et al., 2009]. Chance constrained algorithms and, particularly, ambiguous chance constrained algorithms, are instances of robust algorithms. Robust optimization algorithms have been proposed as a means for ensuring that machine learning algorithms, ranging from supervised learning algorithms [Provost and Fawcett, 2001] to reinforcement learning algorithms [García and Fernández, 2015], are in some way "safe" to use. However, there are several reasons why robust optimization is not a remedy to our problem—why robust methods still require the user of the algorithm to have detailed knowledge of the problem at hand in order to preclude undesirable behaviors. For example, most (but not all) robust optimization algorithms perform poorly if the user is unable to define a small set of environments that contains the true environment with high probability—they work best when the user has detailed knowledge about the problem at hand.

However, the most significant limitation of robust optimization methods as a means to preclude undesirable behaviors is that robust optimization does not provide a solution to the problem of multiple objectives. If the user wants to preclude a behavior, then he or she might try to encode the behavior in a soft constraint and then apply a robust algorithm. The problem with this approach is the same as the problem with soft constraints in general: the user must have detailed knowledge about the problem in order to properly select the parameter, $\lambda$, that trades off between the primary objective and the soft constraint. Alternatively, the user might treat the objective function and the constraint on behavior as two separate objectives, and apply a robust optimization algorithm designed for multiobjective problems [Deb and Gupta, 2006]. However, these algorithms are still based on Pareto frontiers, and so to select a solution from within the Pareto frontier, as discussed in §4.6, the user must still effectively choose a value for the parameter $\lambda$.

Similarly, *risk-sensitive* optimization methods (which are sometimes viewed as a type of robust optimization) use an objective function, $f$, that captures different measures of the distribution of outcomes that can result from using a solution, $\theta$. For example, if each $\theta \in \Theta$ defines a different controller for a robot, then $f(\theta)$ is often chosen to be a measure of the expected performance of the robot if it uses the controller $\theta$. Risk-sensitive methods might define $f(\theta)$ to also incorporate a term that penalizes solutions, $\theta$, that cause the observed performance of the robot to have high variance [Kuindersma et al., 2012]. Alternatively, risk-sensitive methods might define $f(\theta)$ to be the *expected shortfall* or *conditional value at risk* (CVaR) of the solution $\theta$—a measure of the expected performance of the robot during the worst trials, when using the controller $\theta$ [Tamar et al., 2014; Chow and Ghavamzadeh, 2014]. Although risk-sensitive methods typically do not require detailed problem-specific knowledge to apply, they do not allow the user to specify undesirable behaviors from within a broad class of possible behaviors. Furthermore, risk-sensitive methods typically do not provide practical guarantees about the performances of the solutions that they return.

# 5 A New framework for Designing Machine Learning Algorithms

So far we have discussed the standard approach for designing machine learning algorithms, and its limitations. We now turn to defining a new framework for designing machine learning algorithms that allows the user to, without detailed knowledge of the problem at hand, define and regulate desirable and undesirable behaviors. The first step in our framework is not to define mathematically the goals of the *algorithm* (to find a good solution), but to define mathematically the goals of the *researcher* designing the algorithm (to find a good algorithm). Thus, the crux of our framework is a new mathematical problem formalization, which we call a *Seldonian optimization problem* (SOP) as an homage to Isaac Asimov's fictional character, Hari Seldon, a resident of a universe where Asimov's three (non-probabilistic) laws of robotics failed to adequately control agent behavior, and who formulated and solved a machine learning problem that would likely have required some form of probabilistic constraints [Asimov, 1951]. Before defining a Seldonian optimization problem formally, we discuss the drawbacks of two potential alternatives that are flawed, but which might at first seem more reasonable.

## 5.1 Potential New Approach: Add Constraints on the Probability that a Solution is Safe

Intuitively, we would like to specify that an algorithm should guarantee that, with high probability, undesirable behaviors will not occur. Let $\mathcal{S} \subseteq \Theta$ be the set of *safe solutions*—the solutions that do not cause undesirable behavior or that cause the probability of undesirable behavior to be sufficiently small. For example, $\mathcal{S}$ could be the set of artificial neural networks that, if used to control a robot, cause the probability that the robot will harm a human to be less than some user-defined threshold. We might provide the user with a language for specifying $\mathcal{S}$ and then use a problem formulation of the form:

$$\theta^\star \in \arg\max_{\theta \in \Theta} f(\theta)$$
$$\text{s.t. } \Pr(\theta \in \mathcal{S}) \geq 1 - \delta,$$

where $\delta \in [0, 1]$ is some user-specified confidence level. The problem with this approach is that it is not meaningful to reason about the probability that a particular solution, $\theta$, is in $\mathcal{S}$. Either $\theta \in \mathcal{S}$ or $\theta \notin \mathcal{S}$, and so $\Pr(\theta \in \mathcal{S})$ is necessarily either 0 or 1, and the above problem formulation is equivalent to

$$\theta^\star \in \arg\max_{\theta \in \mathcal{S}} f(\theta),$$

which suffers from the problem that we do not know $\mathcal{S}$ *a priori*, as discussed in §4.4.

## 5.2 Potential New Approach: Add Constraints on the Agent's Data-Driven Belief that a Solution is Safe

One way to avoid the problem described in the previous subsection is to include constraints that are based on the agent's data-driven beliefs about whether $\theta \in \mathcal{S}$, rather than the probability that $\theta \in \mathcal{S}$. That is, we might use a problem formulation of the form:

$$\theta^\star \in \arg\max_{\theta \in \Theta} f(\theta)$$
$$\text{s.t. } \mathcal{B}(\theta \in \mathcal{S} | D) \geq 1 - \delta,$$

where $\mathcal{B}(\theta \in \mathcal{S}|D)$ denotes a measure of the agent's belief that $\theta \in \mathcal{S}$ given data $D$. For example, one might define $\mathcal{B}$ to be the *likelihood function*, $\mathcal{L}(\theta \in \mathcal{S}|D) = \Pr(D|\theta \in \mathcal{S})$ [Bishop, 2006], or perhaps $\mathcal{B}(\theta \in \mathcal{S}|D) \coloneqq 1 - \mathcal{L}(\theta \notin \mathcal{S}|D)$. This approach has two problems. First, the feasible set, $\{\theta \in \Theta : \mathcal{B}(\theta \in \mathcal{S}|D) \geq 1-\delta\}$, is a function of the data, which is a random variable, and so the feasible set is a random variable. Having a problem formulation where the feasible set is a random variable suggests that the problem formulation does not accurately capture our goals when designing an algorithm, since our goals are not random.

The second problem with this approach is more damning: this problem formulation could result in unsafe solutions with high probability, even when there is only a single constraint. That is, if we view $\theta^\star$ as a function of the training data, $D$, the probability that $\theta^\star(D)$ is not in $\mathcal{S}$ could be large. It is straightforward to show that (as a consequence of Boole's inequality) if using either definition of $\mathcal{B}$ proposed above or other definitions of $\mathcal{B}$ based on concentration inequalities, the probability that $\theta^\star(D) \notin \mathcal{S}$ can be as large as $\min\{1, \delta|\Theta|\}$, which will be large for problems where $|\Theta|$ is large. To avoid this, the constraint that we introduce should not require that the feasible set only include solutions that are believed to be safe—it should require that the solution output by the algorithm is safe with high probability. This is what a Seldonian optimization problem—the problem formulation used by our framework—does.

## 5.3 Seldonian Optimization Problem (SOP)

Under our approach, the first step is to define mathematically the goal of the researcher creating the machine learning algorithm. This results in a problem formulation that describes a search over algorithms, rather than solutions, with constraints over the set of algorithms, not over the set of solutions. This means that the constraints (now over algorithms) can constrain the probability that the algorithm will return an unsafe solution. For simplicity, here we focus on the *batch* setting, where an algorithm is given all of the available data at once, as opposed to the *online* setting where data arrives over time [Abu-Mostafa et al., 2012]. To precisely define our new problem formalization, which we call a *Seldonian optimization problem* (SOP), we begin by defining a few additional terms.

A *machine learning algorithm*, $a$, is a function that takes data as input and produces as output a solution. Let $\mathcal{D}$ be the set of all possible data sets that could be given as input to the algorithm, and let $D \in \mathcal{D}$ be a random variable that represents all of the data given to the algorithm. Let $\mathcal{A}$ be the set of all possible machine learning algorithms, each of which is a function, $a : \mathcal{D} \to \Theta$. This definition of machine learning algorithms allows for nondeterministic algorithms since the data set can be defined to contain any random numbers required by the algorithm.

Whereas in the standard approach the objective function takes solutions as input, in a SOP the search is over algorithms, and so the objective function takes algorithms as input. That is, $f : \mathcal{A} \to \mathbb{R}$. A SOP also allows for $n \in \mathbb{N}_{\geq 0}$ *behavioral constraints*, which are constraints on the set of algorithms. Each constraint, indexed using $i \in \{1, \ldots, n\}$, has two components, a *constraint objective*, $g_i : \Theta \to \mathbb{R}$, and a *constraint confidence level*, $\delta_i \in [0, 1]$. The designer of the algorithm must ensure that the algorithm he or she creates satisfies the inequalities: $\Pr(g_i(a(D)) \leq 0) \geq 1 - \delta_i$, for all $i \in \{1, \ldots, n\}$. That is, if the user defines $g_i$ such that $g_i(\theta)$ being greater than zero means that undesirable behavior has occurred, then an algorithm that satisfies the behavioral constraints will ensure that with high probability (where the user can specify the necessary probability) it will not return a solution that causes undesirable behavior.

In summary, a SOP can be written as:

$$\arg\max_{a \in \mathcal{A}} f(a) \qquad (5)$$
$$\text{s.t. } \forall i \in \{1, \ldots, n\}, \Pr(g_i(a(D)) \leq 0) \geq 1 - \delta_i.$$

Here the designer of the algorithm selects the objective function, $f$, the set of algorithms considered, $\mathcal{A}$, what the data, $D$, will include, and a class of allowable constraint objective functions, $g_i$. Once a machine learning algorithm has been designed using this new framework, the user of the algorithm can apply it by selecting specific definitions of $g_i$ from within the class chosen by the researcher who designed the algorithm, and by selecting the desired confidence levels, $\delta_i$. In some cases, the user might still be given additional freedom to select other terms, like the feasible set, $\Theta$, or objective function, $f$.

Notice that, when defining the constraints for a SOP, the term $\forall i \in \{1, \ldots, n\}$ is outside of the $\Pr(\cdot)$ term. This means that each behavioral constraint must hold independently with its associated probability, $1 - \delta_i$. Joint constraints—two different constraints, $g'$ and $g''$, that must both hold simultaneously with some probability, $\delta$—can be encoded within a single constraint in the SOP. For example, one could define a single constraint function, $g_i$, that encodes that both $g'$ and $g''$ should hold: $g_i(x) \coloneqq 0$ if both $g'(x) \leq 0$ and $g''(x) \leq 0$, and $g_i(x) \coloneqq 1$ otherwise.

Consider how a linear regression algorithm that is a solution to a SOP could be applied to our illustrative example from §4.1. The researcher designing the linear regression algorithm would select $f$ to be an objective function like the mean squared error, $\Theta$ to be the set of all possible linear functions, and $D$ to be the training set as described previously. To apply the algorithm, the user creating an agent need only specify $g_i$ and $\delta_i$. For our illustrative example, the user might choose to use a single behavioral constraint, with $g_1(a(D)) \coloneqq |d(a(D))| - \epsilon$, so as to guarantee that, with probability at least $1 - \delta_1$, the absolute value of the discriminatory statistic will be at most $\epsilon$, where $\epsilon$ and $\delta_1$ are chosen by the user.

Notice that to apply the algorithm, the user need not know whether any particular line, $\theta \in \Theta$, causes $g_1(\theta) \leq 0$—it is left to the algorithm to reason about this. That is, the user of the algorithm need not know the value of $g_i(\theta)$ for any particular $\theta$. Instead, the user need only be able to specify $g_i$. As exemplified by our later reinforcement learning example, this means that the user need only be able to recognize undesirable behavior, not which solutions cause undesirable behaviors. The job of analyzing the available data to understand the problem at hand well enough to satisfy the constraint is placed entirely on the machine learning algorithm. Furthermore, notice that although a SOP is more complicated than the problem formulation used in the standard approach, and the job of the researcher using our framework to design a machine learning algorithm might be more difficult, the job of the user who wishes to constrain the behavior of an agent has been drastically simplified (by removing the need for the user to perform data analysis).

## 5.4 Optimal Algorithmic Solutions and Seldonian Algorithms

A SOP, as defined in (5), provides a principled way for specifying constraints on an algorithm. However, it has a few drawbacks that we discuss here and in the following subsections. First, finding an *optimal algorithmic solution*, $a^\star$, which is an algorithm that maximizes $f$ subject to the behavioral constraints, is generally intractable. However, the researcher designing an algorithm should at least ensure that the algorithms that he or she proposes satisfy the behavioral constraints. We refer to such an algorithm—an algorithm that satisfies all of the behavioral constraints—as a *Seldonian algorithm*.

## 5.5 Quasi-Seldonian Algorithms

Although the definition of a SOP captures what we want in an algorithm, for some small-data applications it may not be practical—it may not be possible to find a Seldonian algorithm, let alone an optimal algorithmic solution. Our ability to ensure that an algorithm satisfies a behavioral constraint typically requires the use of *concentration inequalities* [Massart, 2007] like Hoeffding's inequality [Hoeffding, 1963]. Because the confidence bounds produced by these inequalities hold for *any* distribution (given minor verifiable assumptions), they tend to be overly conservative. This means that all Seldonian algorithms for some problems might require an impractical amount of data before the confidence levels required by the behavioral constraints can be satisfied.

To remedy this, we propose *quasi-Seldonian algorithms*, which are algorithms that satisfy the behavioral constraints of a SOP using approximate concentration bounds. For example, rather than using Hoeffding's inequality, one might use Student's $t$-test or a bootstrap confidence bound [Efron, 1987]. These methods produce bounds that are often much tighter than those of Hoeffding's inequality, which means that quasi-Seldonian algorithms can be created for problems where there is insufficient data to find viable Seldonian algorithms. The drawback of these approximate concentration bounds is that they rely on assumptions that may not hold for the problem at hand (such as that the sum of a large number of random variables is normally distributed), but which are often reasonable (for example, the sum of a large number of random variables is *approximately* normally distributed due to the central limit theorem), which means that the resulting confidence bounds only approximately hold—they are not real confidence bounds. Despite their drawbacks, the use of approximate concentration bounds like Student's $t$-test and bootstrap confidence bounds remains commonplace in the sciences, including high-risk medical research [Chambless et al., 2003; Folsom et al., 2003].

One limitation of quasi-Seldonian algorithms is that the user of the algorithm might not know what false assumptions the algorithm makes—a researcher could propose an algorithm with egregious false assumptions that would make it irresponsible to use the algorithm for many applications. To remedy this, researchers should clearly state the assumptions made by quasi-Seldonian algorithms, and should strive to ensure that for most problems the quasi-Seldonian algorithm ensures that $\forall i \in \{1, \ldots, n\}$, $\Pr(g_i(a(D)) \leq 0) \gtrsim 1 - \delta_i$. That is, the probability that a solution is returned that causes undesirable behavior should be less than or *approximately equal to* $\delta_i$.

## 5.6 Existence of (Quasi-)Seldonian Algorithms

Even using approximate confidence bounds, there may not be sufficient data for any algorithm to ensure that the behavioral constraints are met. Rather than require an algorithm to still produce a solution (in which case the algorithm may not be able to remain Seldonian), we propose allowing the algorithm to return NO SOLUTION FOUND (NSF). The algorithm should return NSF when it is unable to find a satisfactory solution given the available data.

To add NSF to the problem formulation, one need only define $\Theta$ such that $\text{NSF} \in \Theta$ and $g_i$ such that $g_i(\text{NSF}) > 0$ for all $i$. This modification is sufficient to ensure that Seldonian algorithms exist. However, it also presents a limitation of our framework: it is most applicable to problems where it is permissible for the algorithm to return NSF. This is usually the case for applications where some existing mechanism (solution), $\theta_0$, is in place for making decisions, and the machine learning algorithm is used to improve upon this mechanism.[1] We do not assume that $\theta_0$, called the *baseline solution*, is in the feasible set, $\Theta$. For these applications, if a Seldonian algorithm returns NSF, we can revert to using the baseline solution. Still, for some applications, a decision other than NSF must be made, and so the Seldonian problem formulation may not be viable.

Notice that, if NSF is included in the SOP, then the algorithm that always returns NSF is Seldonian. This does not mean that including NSF means that a researcher designing a machine learning algorithm should propose this algorithm. If the Seldonian optimization problem is properly designed, the objective function, $f$, should assign a low utility to this trivial algorithm, and so the researcher designing the algorithm should attempt to find a better algorithm—a Seldonian algorithm that returns NSF less frequently.

---

[1]Petrik et al. [2016] discuss a similar setting, wherein improvements must be ensured, but the algorithm is free to fall back to an existing prior solution.

# 6 Example: A Seldonian Regression Algorithm and its Application

So far we have discussed what our framework for designing machine learning algorithms is. We now turn to developing a few algorithms using our framework in order to show its viability. We begin by considering the problem of designing a Seldonian linear regression algorithm. Recall that designing a Seldonian linear regression algorithm will be more difficult than designing a linear regression algorithm using the standard framework for designing algorithms, but that it should be easier (i.e., not require data analysis) for the user to constrain an agent's behavior when using the Seldonian algorithm.

We assume that there is a real-valued random variable, $Y$, that we would like to predict from a real-vector-valued random variable, $X$. That is, each value, $x \in \mathbb{R}^l$, that $X$ can take is a vector of $l \in \mathbb{N}_{>0}$ real-valued features that we will use to estimate the value of $Y$. We restrict the class of estimators that we consider to linear functions, so $\Theta := \mathbb{R}^l$ and each $\theta \in \Theta$ defines a linear combination of the features, $x \in \mathbb{R}^l$, as $\hat{y}(x, \theta) := \theta^\intercal x$. In order to select the weights, $\theta$, that cause $\hat{y}(X, \theta)$ to be a good estimate of $Y$, we are given $m \in \mathbb{N}_{>0}$ realizations of $(X, Y)$. Each realization also comes with a sample of another random variable, $T \in \{0, 1\}$, that encodes the *type* (e.g., gender) associated with the point $(X, Y)$. So, the training data is a random variable, $D = \{(X_i, Y_i, T_i)\}_{i=1}^m$, where some joint distribution over $(X, Y, T)$ exists but is not known. Notice that although $X_i$, $Y_i$, and $T_i$ are all dependent random variables, the tuples $(X_i, Y_i, T_i)$ and $(X_j, Y_j, T_j)$ are independent and identically distributed for all $i \neq j$.

Given this setup, we can write a SOP that defines our goals when designing the regression algorithm:

$$\arg\max_{a \in \mathcal{A}} -\mathbf{E}\left[(\hat{y}(X, a(D)) - Y)^2\right] \qquad (6)$$
$$\text{s.t. } \forall i \in \{1, \ldots, n\} \Pr(g_i(a(D)) \leq 0) \geq 1 - \delta_i,$$

Importantly, we should allow the user of the algorithm to specify the functions, $g_i$, without knowing their values, $g_i(\theta)$, for any particular solution, $\theta \in \Theta$. Furthermore, the language provided to the user for specifying $g_i$ should be as simple as possible, while encompassing a broad spectrum of possible functions. Additionally, notice that here we have defined a single joint distribution over $X, Y$, and $T$—a single regression problem. If our goal is to produce a regression algorithm that can be applied to many different regression problems, then this same SOP could be defined for a distribution over regression problems—a distribution over joint distributions for $X, Y$, and $T$.

Below we present three algorithms—one Seldonian and two quasi-Seldonian—which are likely not optimal algorithmic solutions, but which satisfy the behavioral constraints and try to maximize the objective function. Although the algorithms that we present can be extended to allow for a broad class of definitions of $g_i$—to allow the user to define his or her own notion of undesirable behavior—we initially restrict our discussion to a single behavioral constraint with the constraint objective:

$$g(\theta) \coloneqq |d(\theta)| - \epsilon = \left| \mathbf{E}\Big[\hat{y}(X,\theta) - Y \Big| T = 0\Big] - \mathbf{E}\Big[\hat{y}(X,\theta) - Y \Big| T = 1\Big] \right| - \epsilon.$$

Later we present a more general quasi-Seldonian regression algorithm and derive a quasi-Seldonian reinforcement learning algorithm that showcases how an algorithm can both allow the user to specify $g_i$ to encode his or her own definition of undesirable behavior.

### 6.1 Non-Discriminatory Linear Regression

First, consider the Seldonian algorithm *non-discriminatory linear regression* (NDLR, pronounced "endler"), presented in Algorithm 5. This algorithm has three main steps. First, the data set, $D$, is partitioned into two sets, $D_1$ and $D_2$, for reasons that we discuss later. Second, the first set, $D_1$, is used to select a single solution, called the *candidate solution*, $\theta_c$, that the algorithm considers returning. Third, the algorithm runs a safety test using $D_2$ to determine whether returning the candidate solution would violate a behavioral constraint. More precisely, the algorithm computes a high confidence upper bound on the absolute value of the discriminatory statistic of the candidate solution—a high confidence upper bound on $|d(\theta_c)|$. If this upper bound is less than $\epsilon$, then the candidate solution is returned, and if it is not, then the algorithm returns NO SOLUTION FOUND. Next we discuss in more detail the two main steps of the algorithm—the safety test and the selection of a candidate solution.

The safety test (lines 4–6 of Algorithm 5) is the component of the algorithm that ensures that it is Seldonian. Regardless of which candidate solution, $\theta_c$, is chosen, this step ensures that the behavioral constraint is satisfied—it ensures that with high probability a line with absolute discriminatory statistic larger than $\epsilon$ is not returned. The key component of this step is a method for producing high confidence upper bounds on the absolute discriminatory statistic given the data set $D_2$. To compute these upper bounds, we use the method `HoeffdingDiscrimUpperBound`, which relies on one form of Hoeffding's inequality [Hoeffding, 1963], which is implemented in `HoeffdingUpperBound`.

The key observation within `HoeffdingDiscrimUpperBound` is that the absolute discriminatory statistic includes an absolute value and so Hoeffding's inequality is not immediately applicable. To remedy this, we observe that bounding the absolute discriminatory statistic (with the absolute value in its definition) to be below $\epsilon$ is equivalent to bounding the discriminatory statistic (without the absolute value) to be between $-\epsilon$ and $\epsilon$. Thus, we replace the single behavioral constraint that the absolute discriminatory statistic is less than $\epsilon$ with two behavioral constraints that ensure that $d(a(D)) \geq -\epsilon$ and $d(a(D)) \leq \epsilon$ with high probability. Furthermore, since these two constraints must hold simultaneously with probability $1 - \delta$, we require each to hold with probability $1 - \delta/2$.

Next consider the selection of the candidate solution, $\theta_c$, in more detail (line 3 of Algorithm 5). Poor choices of $\theta_c$ will result in the algorithm returning NO SOLUTION FOUND. Good choices of $\theta_c$ will be solutions that will be returned by the third step of NDLR—solutions whose discriminatory statistics will be successfully bounded below $\epsilon$ when using $D_2$. Furthermore, good choices of $\theta_c$ should also minimize the mean squared error of their predictions—the goal is not just to satisfy the behavioral constraint (avoid discrimination with high probability) but also to optimize the objective (minimize mean squared error). So, Algorithm 5 selects the candidate solution that it predicts (using $D_1$) will have the lowest mean squared error subject to the constraint that it is predicted that the candidate solution will be returned. To accomplish this optimization, Algorithm 5 relies on the function `HoeffdingCandidateObjective`, which uses a boundary function to constrain the search to solutions that are predicted to be returned in the third step.

Notice that the selection of the candidate policy and the safety test use different (statistically independent) data sets, $D_1$ and $D_2$. This is necessary to avoid the multiple comparisons problem [Miller, 2012] when computing the high confidence bounds on the discriminatory statistic using Hoeffding's inequality. Specifically, without the partitioning of data, Hoeffding's inequality would be applied to random variables that are not statistically independent. Intuitively, partitioning the data ensures that the candidate solution has no knowledge of the data that will be used to test its safety, since access to this data could be used to bias the results of the safety test.

**Algorithm 1:** `HoeffdingUpperBound`$(Z, b, \delta)$: Apply Hoeffding's inequality to upper bound the mean of a random variable. $Z = \{Z_1, \ldots, Z_m\}$ is the vector of samples of the random variable, $b \in \mathbb{R}_{\geq 0}$ is the range of the random variable, and $\delta \in (0, 1)$ is the confidence level of the upper bound.

**1 return** $\frac{1}{m} \left( \sum_{i=1}^{m} Z_i \right) + b\sqrt{\frac{\ln(1/\delta)}{2m}}$;

---

**Algorithm 2:** `PredictHoeffdingUpperBound`$(Z, b, \delta, k)$: Predict what `HoeffdingUpperBound`$(Z', b, \delta)$ would return if given a new array of samples, $Z'$, with $k$ elements. We use a conservative prediction (an over-prediction) because under-predictions can cause NDLR to frequently return NO SOLUTION FOUND when it could safely return a solution.

**1 return** $\frac{1}{m} \left( \sum_{i=1}^{m} Z_i \right) + b\sqrt{\frac{\ln(1/\delta)}{k}}$;

---

**Algorithm 3:** `HoeffdingCandidateObjective`$(D, \delta, \epsilon, b, k)$: The objective function maximized by the candidate solution when using Hoeffding's inequality. Here $k$ is the number of data points that will be used in the subsequent safety test in NDLR—the cardinality of $D_2$.

**1 Input: 1)** Data set $D = \{(X_i, Y_i, T_i)\}_{i=1}^{m}$, **2)** confidence level, $\delta \in (0, 1)$, **3)** maximum level of discrimination, $\epsilon \in \mathbb{R}_{>0}$, **4)** an upper bound, $b$, on the range of possible prediction errors, **5)** a number of samples, $k$.
**2** $m_1 \leftarrow \sum_{i=1}^{m} T_i, \quad m_0 \leftarrow m - m_1$;
**3** Let $(X_i^0, Y_i^0)$ be the $i^{\text{th}}$ data point of type zero, and $(X_i^1, Y_i^1)$ be the $i^{\text{th}}$ data point of type one;
**4** Create array $Z$ of length $\min\{m_0, m_1\}$, with values
$Z_i = (\theta^\intercal X_i^0 - Y_i^0) - (\theta^\intercal X_i^1 - Y_i^1)$;
**5** ub $= \max \Big\{$ `PredictHoeffdingUpperBound`$(Z, b, \delta/2, k)$,
    `PredictHoeffdingUpperBound`$(-Z, b, \delta/2, k) \Big\}$
**6 if** ub $\leq \epsilon$ **then**
**7**     **return** $\frac{1}{m} \sum_{i=1}^{m} (\theta^\intercal X_i - Y_i)^2$
**8 return** $b^2 + \text{ub} - \epsilon$

**Algorithm 4:** `HoeffdingDiscrimUpperBound`$(\theta, D, \delta, \epsilon, b)$: Compute a high-probability upper bound on the absolute discriminatory statistic using Hoeffding's inequality.

---

1 **Input: 1)** Solution, $\theta$, **2)** data set $D = \{(X_i, Y_i, T_i)\}_{i=1}^m$, **3)** confidence level, $\delta \in (0, 1)$, **4)** maximum level of discrimination, $\epsilon \in \mathbb{R}_{>0}$, **5)** an upper bound, $b$, on the range of possible prediction errors.
2 $m_1 \leftarrow \sum_{i=1}^m T_i, \qquad m_0 \leftarrow m - m_1$;
3 Let $(X_i^0, Y_i^0)$ be the $i^{\text{th}}$ data point of type zero, and $(X_i^1, Y_i^1)$ be the $i^{\text{th}}$ data point of type one;
4 Create array $Z$ of length $\min\{m_0, m_1\}$, with values $Z_i = (\theta^\intercal X_i^0 - Y_i^0) - (\theta^\intercal X_i^1 - Y_i^1)$;
5 **return** $\max\{$`HoeffdingUpperBound`$(Z, b, \delta/2)$, `HoeffdingUpperBound`$(-Z, b, \delta/2)\}$

---

**Algorithm 5:** Non-Discriminatory Linear Regression (NDLR, "endler")

---

1 **Input: 1)** Data set $D = \{(X_i, Y_i, T_i)\}_{i=1}^m$, **2)** confidence level, $\delta \in (0, 1)$, **3)** maximum level of discrimination, $\epsilon \in \mathbb{R}_{>0}$, **4)** an upper bound, $b$, on the magnitude of a prediction error.
2 Partition $D$ into $D_1$ (20% of data) and $D_2$ (80% of data);
3 $\theta_c \in \arg\min_{\theta \in \Theta}$ `HoeffdingCandidateObjective`$(D_1, \delta, \epsilon, b, |D_2|)$;
4 **if** *HoeffdingDiscrimUpperBound*$(\theta_c, D_2, \delta, \epsilon, b) \leq \epsilon$ **then**
5     **return** $\theta_c$;
6 **return** NO SOLUTION FOUND;

## 6.2 Quasi-Non-Discriminatory Linear Regression

Next, consider the quasi-Seldonian regression algorithms presented in Algorithm 10, and which we call *quasi-non-discriminatory linear regression* (QNDLR, pronounced "kwindler") and QNDLR($\lambda$). These two algorithms are modifications of NDLR to use Student's $t$-test (implemented in `TTestUpperBound`) rather than Hoeffding's inequality when computing high confidence upper bounds.

Both NDLR and QNDLR attempt to minimize the mean squared error while ensuring that (with high probability) the absolute discriminatory statistic is at most $\epsilon$. Given large amounts of data (e.g., as $m \to \infty$), they tend to converge to solutions with absolute discriminatory statistics slightly less than $\epsilon$. However, the real goal is not to produce solutions with discriminatory statistics close to $\epsilon$—the goal is to avoid discrimination as much as possible. That is, this problem is really a multiobjective problem: the ideal solution should simultaneously minimize mean squared error and the absolute discriminatory statistic.

This is an example of a problem for which multiobjective methods can be combined with our new framework. QNDLR($\lambda$) is a modification of QNDLR to allow for a multiobjective approach while maintaining the high-probability guarantees of a quasi-Seldonian algorithm. Specifically QLDNR($\lambda$) is a variant of QNDLR that includes a soft constraint parameterized by $\lambda$, as in (4), in the objective function used when selecting the candidate solution. Thus, QNDLR($\lambda$) is a solution to the Seldonian optimization problem in (6), modified so that the mean squared error objective includes a penalty proportional to the sample absolute discriminatory statistic. Thus, QNDLR($\lambda$) is a quasi-Seldonian algorithm that ensures with high probability that the absolute discriminatory statistic will be at most $\epsilon$, and subject to this constraint, it simultaneously optimizes mean squared error and discrimination.

Recall that selecting $\lambda$ is a difficult process that often requires additional data analysis, and that values of $\lambda$ that are too small will result in a standard soft-constrained method producing a solution with absolute discriminatory statistic larger than $\epsilon$. Crucially, because QNDLR($\lambda$) is a quasi-Seldonian algorithm, even if the user selects a $\lambda$ that would result in absolute discriminatory statistics larger than $\epsilon$ if using a standard soft-constrained method, QNDLR($\lambda$) will (with high probability) not return a solution with absolute discriminatory statistic larger than $\epsilon$. This effectively eliminates the risk associated with selecting a poor value for $\lambda$, which would make standard soft-constrained methods unsafe, thereby allowing the user to select any value for $\lambda$ that they think properly balances the trade-off between mean squared error and discrimination.

Lastly, notice that NDLR, QNDLR, and QNDLR($\lambda$) have significantly higher computational complexity than ordinary least squares linear regression. The primary computational bottleneck in both algorithms occurs when searching for the solution, $\theta_c$, that optimizes the candidate objective function (lines 3 and 4 of Algorithms 5 and 10, respectively). Different choices of methods for performing the search for $\theta_c$ result in different computational complexities. In our experiments we performed the search for $\theta_c$ using gradient descent with error-based termination conditions. Although the higher computational complexity of the (quasi-)Seldonian algorithms that we present was not an issue in our experiments, for big-data applications it could be a concern. It remains an important open question whether the search for a candidate solution can (perhaps under certain conditions) be further optimized computationally.

---

**Algorithm 6:** `TTestUpperBound`($Z, \delta$): Apply Student's $t$-test to a vector of samples of a random variable. $Z = \{Z_1, \ldots, Z_m\}$ is the vector of samples and $\delta \in (0,1)$ is the confidence level.

---

**1** $\bar{Z} \leftarrow \frac{1}{m} \sum_{i=1}^{m} Z_i, \qquad \sigma \leftarrow \sqrt{\frac{1}{m-1} \sum_{i=1}^{m} (Z_i - \bar{Z})^2}$;

**2 return** $\bar{Z} + \frac{\sigma}{\sqrt{m}} t_{1-\delta, m-1}$;

**Algorithm 7:** `PredictTTestUpperBound`$(Z, \delta, k)$: Predict what `TTestUpperBound`$(Z, \delta)$ would return if given a new array of samples, $Z$, with $k$ elements, and err on the side of over-estimating. We use a conservative prediction (an over-prediction) because under-predictions can cause NDLR to frequently return NO SOLUTION FOUND.

**1** $\bar{Z} \leftarrow \frac{1}{m} \sum_{i=1}^{m} Z_i, \qquad \sigma \leftarrow \sqrt{\frac{1}{m-1} \sum_{i=1}^{m}(Z_i - \bar{Z})^2}$;
**2 return** $\bar{Z} + 2 \frac{\sigma}{\sqrt{k}} t_{1-\delta, k-1}$;

---

**Algorithm 8:** `TTestCandidateObjective`$(D, \delta, \epsilon, k, \lambda)$: The objective function maximized by the candidate solution.

**1 Input: 1)** Data set $D = \{(X_i, Y_i, T_i)\}_{i=1}^{m}$, **2)** confidence level, $\delta \in (0, 1)$, **3)** maximum level of discrimination, $\epsilon \in \mathbb{R}_{>0}$, **4)** a number of samples, $k$, **5)** a constant, $\lambda \in \mathbb{R}_{\geq 0}$ that balances the trade-off between mean squared error and discrimination.
**2** $m_1 \leftarrow \sum_{i=1}^{m} T_i, \qquad m_0 \leftarrow m - m_1$;
**3** Let $(X_i^0, Y_i^0)$ be the $i^{\text{th}}$ data point of type zero, and $(X_i^1, Y_i^1)$ be the $i^{\text{th}}$ data point of type one;
**4** Create array $Z$ of length $\min\{m_0, m_1\}$, with values
$Z_i = (\theta^\intercal X_i^0 - Y_i^0) - (\theta^\intercal X_i^1 - Y_i^1)$;
**5** $\text{ub} = \max\Big\{\text{PredictTTestUpperBound}(Z, b, \delta/2, k),$
$\qquad\qquad \text{PredictTTestUpperBound}(-Z, b, \delta/2, k)\Big\}$
**6 if** $\text{ub} \leq \epsilon$ **then**
**7** $\quad$ **return** $\frac{1}{m} \sum_{i=1}^{m}(\theta^\intercal X_i - Y_i)^2 + \lambda \frac{1}{|Z|} \sum_{i=1}^{|Z|} Z_i$
**8 return** $b^2 + \text{ub} + (\lambda - 1)\epsilon$

---

**Algorithm 9:** `TTestDiscrimUpperBound`$(\theta, D, \delta, \epsilon)$: Compute a high-probability upper bound on the absolute discriminatory statistic using Student's $t$-test.

**1 Input: 1)** Solution, $\theta$, **2)** data set $D = \{(X_i, Y_i, T_i)\}_{i=1}^{m}$, **3)** confidence level, $\delta \in (0, 1)$, **4)** maximum level of discrimination, $\epsilon \in \mathbb{R}_{>0}$.
**2** $m_1 \leftarrow \sum_{i=1}^{m} T_i, \qquad m_0 \leftarrow m - m_1$;
**3** Let $(X_i^0, Y_i^0)$ be the $i^{\text{th}}$ data point of type zero, and $(X_i^1, Y_i^1)$ be the $i^{\text{th}}$ data point of type one;
**4** Create array $Z$ of length $\min\{m_0, m_1\}$, with values
$Z_i = (\theta^\intercal X_i^0 - Y_i^0) - (\theta^\intercal X_i^1 - Y_i^1)$;
**5 return**
$\max\{\text{TTestUpperBound}(Z, \delta/2), \text{TTestUpperBound}(-Z, \delta/2)\}$;

**Algorithm 10:** Quasi-Non-Discriminatory Linear Regression (QNDLR, "kwindler") if $\lambda = 0$ and QNDLR($\lambda$) if $\lambda > 0$.

1 **Assumptions:** This quasi-Seldonian algorithm assumes that the sample discriminatory statistic computed using all of the data is normally distributed, and also uses the minimum mean squared error estimator of variance within Student's $t$-test rather than the unbiased estimator.
2 **Input: 1)** Data set $D = \{(X_i, Y_i, T_i)\}_{i=1}^m$, **2)** confidence level, $\delta \in (0, 1)$, **3)** maximum level of discrimination, $\epsilon \in \mathbb{R}_{>0}$, **4)** a hyperparameter $\lambda \in \mathbb{R}_{\geq 0}$.
3 Partition $D$ into $D_1$ (20% of data) and $D_2$ (80% of data);
4 $\theta_c \in \arg\min_{\theta \in \Theta}$ `TTestCandidateObjective`$(D_1, \delta, \epsilon, |D_2|, \lambda)$;
5 **if** `TTestDiscrimUpperBound`$(\theta_c, D_2, \delta, \epsilon) \leq \epsilon$ **then**
6 $\quad$ return $\theta_c$;
7 **return** NO SOLUTION FOUND;

## 6.3 NDLR and QNDLR Discussion

NDLR, QNDLR, and QNDLR($\lambda$) are all instances of a more general algorithm that allows the user to **1)** select any objective function (e.g., mean squared error alone as used by NDLR and QNDLR, or mean squared error with a penalty proportional to the absolute discriminatory statistic as used by QNDLR($\lambda$)), and **2)** bound any statistic for which the user can provide data-based unbiased estimates. That is, the user provides **1)** a function, $\hat{f}$, such that $\hat{f}(\theta, D) \in \mathbb{R}$ is an estimate of the utility of the solution $\theta$, computed using data $D$ and **2)** a function $\hat{g}$, such that $\hat{g}(\theta, D) \in \mathbb{R}^{|D|}$ is a vector of unbiased estimates of the desired behavioral constraint function, $g(\theta)$ (thus, $g(\theta) \coloneqq \mathbf{E}[\hat{g}(\theta, D)]$). This more general quasi-Seldonian algorithm, presented in Algorithm 11, is an approximate solution to the Seldonian optimization problem:

$$\arg\max_{a \in \mathcal{A}} \mathbf{E}\left[\hat{f}(a(D), D')\right]$$
$$\text{s.t. } \Pr(\mathbf{E}_{D'}[\hat{g}(a(D), D')] \leq 0) \geq 1 - \delta,$$

where the $D$ and $D'$ data sets are independent and identically distributed random variables and $\mathbf{E}_{D'}$ denotes that the expected value is taken only over $D'$ (not $D$).

Intuitively, the quasi-Seldonian algorithm presented in Algorithm 11 is similar to QNDLR—it partitions the data set, uses one partition to compute a candidate solution, and uses Student's $t$-test with the second partition to test whether the candidate solution can be returned. In fact, QNDLR is a special case of this algorithm, where $\hat{f}$ is the negative sample mean squared error and $\hat{g}$ is the vector of sample discriminatory statistics denoted by $Z$ in Algorithm 9, minus $\epsilon$. It is straightforward to extend Algorithm 11 to allow for multiple behavioral constraints and to allow for functions, $\hat{g}$, that produce variable numbers of outputs.

Notice that this more general algorithm could be used to bound other notions of discrimination. For example, one could define $\hat{g}$ to be the difference in predictions (rather than prediction errors) in order to require the mean predictions to be similar for people of each type. In this sense, users are free to select the definitions of discrimination that they each desire. Moreover, users need not know the true values of the statistics that they wish to ensure are bounded. As an example of this, notice that specifying the $\hat{g}$ function used by QNDLR does not require knowledge of the true discriminatory statistic of any solutions.

Lastly, consider what would happen if the user asked for the impossible: what if the user desired the difference in mean predictions for people of each type, as well as the difference in mean prediction errors for people of each type, to both be small? For our illustrative example it is straightforward to show that this is not possible. As a result, any Seldonian algorithm, including Algorithm 11, should return NO SOLUTION FOUND with high probability—that is, a (quasi-)Seldonian algorithm can effectively say "I cannot do that."

---

**Algorithm 11:** An example of a general-purpose quasi-Seldonian algorithm. Here $\mu(v)$ denotes the average of the elements of the vector $v$ and $\sigma(v)$ denotes the sample standard deviation of the elements of the vector $v$ including Bessel's correction.

1 **Input** : **1)** Feasible set, $\Theta$, **2)** data set, $D$, **3)** probability, $1 - \delta$, **4)** function, $\hat{f}$, such that $\hat{f}(\theta, D) \in \mathbb{R}$ is an estimate of the utility of the solution $\theta$, computed using data $D$, **5)** function $\hat{g}$, such that $\hat{g}(\theta, D) \in \mathbb{R}^{|D|}$ is a vector of unbiased estimates of $g(\theta)$.
2 **Output:** A solution, $\theta \in \Theta$, or NO SOLUTION FOUND.
3 Partition $D$ into two data sets, $D_1$ and $D_2$;
4 $\theta_c = \arg\max_{\theta \in \Theta} \hat{f}(\theta, D_1)$    s.t.    $\mu(\hat{g}(\theta, D_1)) + 2\frac{\sigma(\hat{g}(\theta, D_1))}{\sqrt{|D_2|}} t_{1-\delta, |D_2|-1} \leq 0$;
5 **if** $\mu(\hat{g}(\theta_c, D_2)) + \frac{\sigma(\hat{g}(\theta_c, D_2))}{\sqrt{|D_2|}} t_{1-\delta, |D_2|-1} \leq 0$ **then**
6 | **return** $\theta_c$;
7 **return** NO SOLUTION FOUND;

---

We now discuss a different point: given small amounts of data, both NDLR and QNDLR may often return NO SOLUTION FOUND. This raises the question: what should one do if faced with a problem where there is not sufficient data to guarantee (even using approximate concentration bounds) that the resulting solution will not produce discriminatory behavior? One tempting solution is to argue that, in this case, one should ignore the behavioral constraints and simply use ordinary least squares linear regression. However, we contend that insufficient data is not an excuse to immediately return to using methods that provide no guarantees about the safety of their behavior.

Instead, one might use algorithms that relax the behavioral constraints even more than quasi-Seldonian algorithms. Intuitively, Seldonian algorithms ensure that the probability of undesirable behavior is at most $\delta$, while quasi-Seldonian algorithms ensure that the probability of undesirable behavior will be less then or approximately equal to $\delta$. One might go a step further and define even weaker variants of quasi-Seldonian algorithms with other properties. For example, one might require the algorithm to produce solutions in a way such that a third party could not show negligence—so that someone else could not use the data available to the algorithm to show with high confidence that the algorithm produced a solution whose discriminatory statistic has magnitude greater than $\epsilon$. However, here we focus primarily on quasi-Seldonian algorithms. Seldonian algorithms like NDLR tend to require too much data to be practical, and while weaker variants of quasi-Seldonian algorithms can produce solutions with any amount of data, they do not provide satisfying guarantees about their behavior. Quasi-Seldonian algorithms like QNDLR provide a nice middle-ground between these two extremes—they can produce solutions given practical amounts of data, and they also provide the user with practical insights into the probability that undesirable behaviors might occur.

## 6.4 Related Work

NDLR, QNDLR, and QNDLR($\lambda$) are not the first supervised learning algorithms that have been proposed as a means to preclude discriminatory behavior or ensure fairness. However, to the best of our knowledge, they are the first to provide the user with a practical guarantee about the probability of undesirable behavior (discrimination).

Kamiran and Calders [2009] provide classification algorithms that attempt to ensure that the probability of a desirable prediction is similar regardless of which type a person is. In the context of linear regression, this would be similar to using a different definition of the discriminatory statistic that considers the difference between the mean predictions for people of each type, rather than the difference between the mean prediction *errors*:

$$d(\theta) \coloneqq \mathbf{E}\left[\hat{y}(X,\theta)|T=0\right] - \mathbf{E}\left[\hat{y}(X,\theta)|T=1\right]. \tag{7}$$

There have been many extensions and adaptations of this work [Calders and Verwer, 2010; Luong et al., 2011; Kamishima et al., 2011; Dwork et al., 2012; Feldman et al., 2015; Fish et al., 2016; Joseph et al., 2016] and a long history of related game-theoretic efforts [Rabin, 1993; Fehr and Schmidt, 1999; Falk and Fischbacher, 2006].

There have also been recent efforts to solve the related problem of determining whether a deployed machine learning algorithm is producing discriminatory behavior (as opposed to designing non-discriminatory algorithms). For example several authors have proposed means for determining how sensitive a black-box supervised learning algorithm is to each feature in a vector of features used to represent the input [Datta et al., 2016; Adler et al., 2016], which can be used to determine the impact of features like race and gender [Datta et al., 2015].

Other researchers have considered methods for precluding forms of discrimination that are similar to the discriminatory statistic used by NDLR and QNDLR (defined in (3)). For example, Hardt et al. [2016] propose a method for adjusting the predictor produced by any standard classification or regression algorithm so that it achieves *equalized odds* and *equality of opportunity*. A predictor, $\hat{y} : \mathcal{X} \to \mathbb{R}$, achieves *equalized odds* if $y(X)$ is statistically independent of $T$ given $Y$, and a binary classifier, $\hat{y} : \mathcal{X} \to \{0, 1\}$ achieves *equality of opportunity* if

$$\Pr\left(\hat{y}(X) = 1 \Big| T = 0, Y = 1\right) = \Pr\left(\hat{y}(X) = 1 \Big| T = 1, Y = 1\right). \qquad (8)$$

The notion of equality of opportunity is particularly similar the notion of discrimination used by NDLR and QNDLR since it bounds differences in mean prediction errors rather than differences in mean predictions. Consider an example where $Y = 1$ denotes that a college applicant would graduate from college, while $Y = 0$ denotes that the applicant would not graduate, and where $T \in \{0, 1\}$ denotes the applicants type (e.g., gender). A predictor of $Y$ that takes an applicant's application, $X$, as input achieves equality of opportunity if, considering only those students who would actually graduate from college, the predictor has the same prediction error for applicants of each type.

The difference between this definition of discrimination and the one used by NDLR and QNDLR (see (3)) is that (3) does not condition on $Y = 1$—on the applicants who would actually graduate from college. This difference becomes clear when we consider a hypothetical predictor that predicts that all male applicants will graduate (regardless of whether or not they really will graduate), but which correctly predicts whether each female applicant will graduate. The only applicants that this predictor predicts will *not* graduate are female applicants who really will not graduate. Using (3) this predictor would be deemed discriminatory because the mean prediction error for male applicants would be roughly 0.5 (if roughly half of applicants would graduate) while the mean prediction error for female applicants would be 0.0. However, this predictor would be deemed non-discriminatory if using the notion of equality of opportunity since, if we only consider the applicants who would actually graduate, the mean prediction errors are the same for males and females (zero in both cases).

However, the primary difference between our work and these previous works is *not* our differing definitions of discrimination. We selected the discriminatory statistic presented in (3) as an example, which we used to create NDLR and QNDLR as simple first examples of (quasi-)Seldonian algorithms. Both NDLR and QNDLR are special cases of the more general (quasi-)Seldonian algorithm presented in Algorithm 11, which allows its users to specify their own definitions of undesirable behavior (in this case, discrimination). That is, Algorithm 11 allows the user to select the definition of discrimination that he or she believes is most suitable for an application. For example, Algorithm 11 can be used to bound any of the aforementioned definitions of discrimination with high probability, including (7) or the difference between the left and right sides of (8). Moreover, Algorithm 11 could be used to bound statistics that are unrelated to discrimination (e.g., the variance of predictions or the percent of all predictions that are above some threshold). This is in stark contrast to the aforementioned methods for algorithmic fairness that each only allow the user to mitigate a single particular form of discrimination.

To further clarify this point, consider again our illustrative example. For this example it is straightforward to verify that there does not exist a single estimator that bounds both the difference in mean predictions (see (7)) and the difference in mean prediction errors (see (3)) to both be small. However, the user of Algorithm 11 *could* define behavioral constraints to require that both of these statistics be bounded below a small constant with high probability. Any (quasi-)Seldonian algorithm, including Algorithm 11, should return NO SOLUTION FOUND with high probability in this case. This is effectively the algorithm's way of stating "I cannot do what was requested." Existing methods for algorithmic fairness do not have this capability because they do not give their users the freedom to select their own desired definitions of undesirable behavior (discrimination) from within a sufficiently broad class.

## 6.5 Applications to the Illustrative Example

We applied SCLR, NDLR, and QNDLR to the illustrative example from §4.1, using $\delta = 0.05$, $\epsilon = 0.1$, and using different amounts, $m$, of training data. The results are summarized in Figure 7. Notice that NDLR is extremely conservative—not once did it return a solution with discriminatory statistic larger than $\epsilon$. The main limitation of NDLR is that it requires a large amount of data (around $m = 500{,}000$) before it begins to return a solution more often than not. QNDLR is not quite as conservative as NDLR, although it still maintains an error probability less than $\delta$, and requires only around $m = 1{,}500$ training points before it begins to return a solution more often than not. This is an example where QNDLR provides a nice balance of the trade-off between data efficiency (the amount of data needed to find solutions) and how much its probabilistic guarantees can be trusted. Notice also that both NDLR and QNDLR produce solutions that have higher mean squared error than the ordinary least squares fit. This is because, as discussed previously, minimizing mean squared error and the absolute discriminatory statistic are at odds—the algorithms must balance the trade-off between error and discrimination.

We also provide empirical comparisons to a few other algorithms. First, we include *soft constrained linear regression* (SCLR), which includes soft constraints as described in §4.5 using various settings of $\lambda$ (we write SCLR($\lambda$) to denote SCLR applied with a specific value of $\lambda$). Notice that larger values of $\lambda$ result in smaller discriminatory statistics, and that the magnitude of the discrminatory statistic is sensitive to the choice of $\lambda$—if it is selected to be too large, the mean squared error is unnecessarily high, while if it is too small, it will result in too much discrimination. Furthermore, notice that even when using the setting of $\lambda$ that results in the mean absolute discriminatory statistic being approximately $\epsilon$, SCLR always returns a solution, and so given small amounts of data it can often produce solutions that discriminate too much.

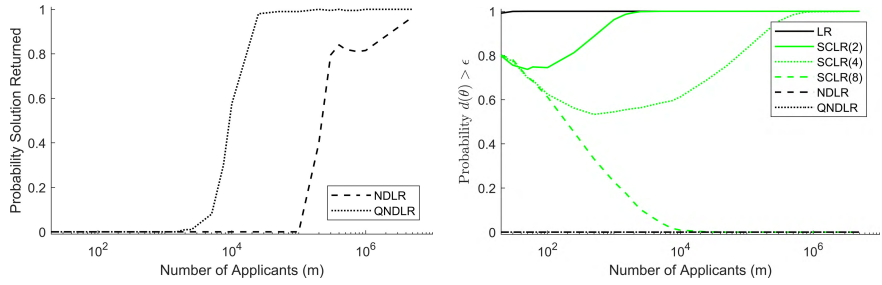

(A) The probability that a solution was returned for various training set sizes, $m$. The curve for QNDLR is covered perfectly by the curve for QNDLR($\lambda$).

(B) The probability that each method will produce a solution that discriminates too much—a $\theta$ where $|d(\theta)| > \epsilon$—for various training set sizes, $m$.

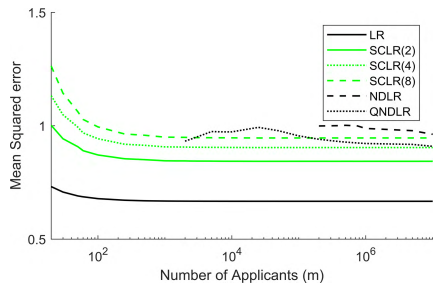
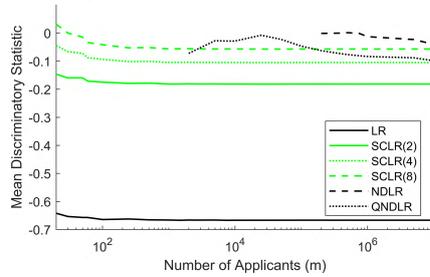

(C) The mean squared error of the solutions produced by each method for various training set sizes, $m$.

(D) The mean (over all 2,000 trials) absolute discriminatory statistic, $|d(\theta)|$, using the solutions produced by each method with various training set sizes, $m$.

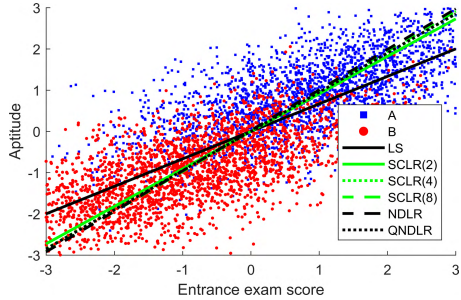

(E) Examples of the lines found by several of the algorithms given $m = 500{,}000$ training points.

Figure 7: Results of applying various linear regression algorithms to the illustrative example of §4.1. All results are averaged over 2,000 trials. LR denotes ordinary least squares linear regression.

## 6.6 Application of NDLR and QNDLR to Real Data

The illustrative example presented in §4.1 provides an easily reproduced and understood example of how discriminatory behavior can manifest when using machine learning algorithms designed using the standard approach, and how it can be mitigated when using (quasi-)Seldonian algorithms. However, it is a simple synthetic example, which raises the question: does this sort of discriminatory (racist or sexist) behavior actually occur when using machine learning algorithms designed using the standard approach with real data? Recent research has suggested that the answer is affirmative: machine learning algorithms designed using the standard approach that were applied to important problems have acted in racist and sexist ways. For example, Datta et al. [2016] showed that standard classification algorithms including logistic regression, support vector machines, and random forests can all produce racist and sexist behavior when used to predict whether or not a person is likely to commit a crime in the future. Similarly, Datta et al. [2015] showed that Google's online advertising system is more likely to show advertisements for a high paying job to men than it is to show it to women, and Kay et al. [2015] found that "image search results for occupations slightly exaggerate gender stereotypes and portray the minority gender for an occupational [sic] less professionally." This real-world discrimination is not limited to gender: Sweeney [2013] found that Google AdSense was more likely to generate advertisements suggestive of an arrest record when searching for names associated with black people when compared to searches for names associated with white people. In another example, machine learning algorithms have been used to predict whether convicts will be likely to commit crimes in the future, and these machine predictions were considered during sentencing. A recent investigation into the behavior of these algorithms suggests that they may be twice as likely to *incorrectly* predict that a black person is likely to commit a crime than they are to incorrectly predict that a white person is likely to commit a crime [Angwin et al., 2016].

In this section we present another example of how regression algorithms designed using the standard approach can be applied to real data, how this results in sexist behavior, and how (quasi-)Seldonian algorithms can be used to preclude sexist behavior. Whereas in our illustrative example we predicted the aptitude of job applicants based on their résumé, here we predict the aptitude of students applying to a university. Specifically, we use applicants' scores on nine exams taken as part of the application process to a university to predict what their *grade-point averages* (GPAs) will be during the first three semesters at university. Our training set consisted of data from 43,303 Brazilian students, which was originally collected to evaluate Brazil's *quota* system [Htun, 2004]. For each student, we are given three terms: $X, Y$, and $T$, as in the illustrative example. Here, $X \in \mathbb{R}^9$ is a vector of the student's 9 exam scores, $Y$ is the student's mean GPA during the first three semesters of university (using the system $A = 4.0$, $B = 3.0$, $C = 2.0$, $D = 1.0$, and $F = 0.0$), and $T$ is a binary value that indicates if the student is female or male.

For these experiments, we used *leave-one-out cross-validation*. That is, we predicted the performance of the $i^{\text{th}}$ student after training using the data from all of the other students, and measured the resulting sample mean squared error and discrminatory statistic when using various regression algorithms. We applied QNDLR and QNDLR($\lambda$) using $\delta = 0.05$ and $\epsilon = 0.05$, as well as four algorithms designed using the standard approach: *least squares linear regression* (LS), an artificial neural network with ten neurons in its hidden layer (ANN) [Demuth and Beale, 1993], a random forest (RF) [Liaw and Wiener, 2002], and *soft-constrained linear regression* (SCLR). For the first three machine learning methods designed using the standard approach (LS, ANN, and RF) we used the standard implementations provided by MATLAB 2015a, and we implemented SCLR by performing gradient descent on the objective function provided in (4). We applied SCLR with $\lambda = 0.1$ and $\lambda = 1.0$, which we denote by SCLR(0.1) and SCLR(1.0) respectively, and we used these same values of $\lambda$ when running QNDLR($\lambda$).

Figure 8 depicts the result of applying LS, ANN, RF, SCLR(0.1), SCLR(1.0), QNDLR, QNDLR(0.1), and QDNRL(1.0). In all trials (all 43,303 folds of leave-one-out cross-validation) QNDLR, QNDLR(0.1), and QNDLR(1.0) all always returned solutions (not NSF). Notice that the algorithms designed using the standard approach discriminate against female applicants (they produce large discriminatory statistics), while QNDLR and QNDLR($\lambda$) do not (they succesfully bound the discriminatory statistic so that it is less that $\epsilon = 0.05$). The sample discriminatory statistics (computed using leave-one-out cross-validation and rounded to two significant figures) when using the standard methods are $-0.28$ (LR), $-0.27$ (ANN), $-0.27$ (RF), $-0.26$ (SCLR(0.1)), and $-0.04$ (SCLR(1.0)), while the sample discriminatory statistics of QNDLR, QNDLR(0.1), and QNDLR(1.0) are 0.03, 0.03, and 0.00, respectively. Although seemingly small numbers, the discriminatory statistics for the standard methods (other than SCLR(1.0), which we discuss later) correspond to massive systematic discrimination against female applicants. To put into context the magnitude of a discriminatory statistic of $-0.27$, consider Figure 9, which provides a histogram of the GPAs of all 43,303 students, and notice that due to the clustering of GPAs towards the upper end of the spectrum, a difference of 0.27 is significant.

Notice that SCLR($\lambda$) is capable of precluding significant discrimination when $\lambda$ is properly tuned. Earlier we argued that selecting appropriate values for $\lambda$ *a priori* (before observing the data from the problem at hand) is difficult. Notice that using $\lambda = 0.1$ results in significant discrimination against female applicants. Furthermore, the standard error bars in Figure 8 indicate that even using $\lambda = 1.0$ resulted in solutions with discriminatory statistics above 0.05 with probability greater than $\delta$. By contrast, QNDLR($\lambda$) combines a soft-constrained multiobjective approach with our framework. Even when using values for $\lambda$ that would produce significant discrimination (e.g., $\lambda = 0.1$), QNDLR($\lambda$) still ensures that the discriminatory statistic will be below $\epsilon$ with probability at least $1 - \delta$. Also, notice that the mean prediction error for male applicants when using QNDLR(0.1) is above $\epsilon/2 = 0.025$. This does *not* mean that QNDLR(0.1) violates the behavioral constraint since the mean error for female applicants is not below $-\epsilon/2 = -0.025$. That is, the difference in mean prediction errors remains well below 0.05.

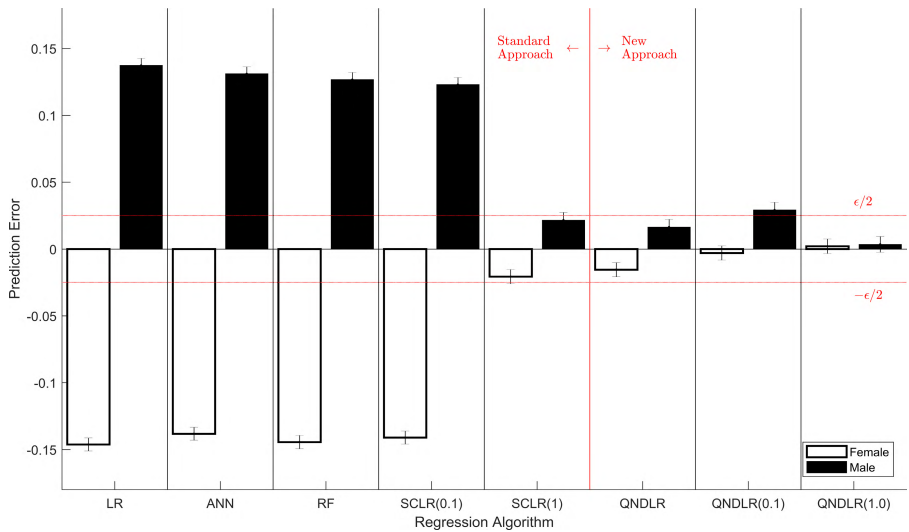

Figure 8: Empirical results using various regression algorithms to predict student GPAs based on entrance exam scores. Results are averaged over all 43,303 folds when using leave-one-out cross-validation, and standard error bars are provided.

A common misconception about Figure 8 is that it shows that QNDLR and QNDLR($\lambda$) produced more accurate predictions for all students—they did not. To emphasize this point, in Figure 10 we provide additional information about the estimators of GPA produced by each method. Notice that, just like in the illustrative example, the non-discriminatory algorithms tend to produce estimates with slightly higher mean squared error than the algorithms designed using the standard approach, but significantly less discrimination. This slightly higher mean squared error is the cost associated with precluding significant discrimination when applying a quasi-Seldonian algorithm for this application.

# 7 Example: A Seldonian Reinforcement Learning Algorithm and its Application to Diabetes Treatment

So far we have focused on (quasi-)Seldonian algorithms that solve a relatively simple problem—fitting a line to data—in order to provide an easily accessible and broadly interesting example. We now turn to showing that the ideas that we have presented are not limited to this simple setting: we can create (quasi-)Seldonian algorithms that tackle more sophisticated problems. Specifically, we propose a (quasi-)Seldonian *batch reinforcement learning* algorithm. Reinforcement learning algorithms allow machines to learn by trial and error without the need for an oracle (or teacher) that tells the algorithm or agent what decisions it should have made, and have been applied to a variety of challenging and real-world problems [Tesauro, 1995; Ng et al., 2004; Mnih et al., 2015]. For an introduction to reinforcement learning, see the work of Sutton and Barto [1998].

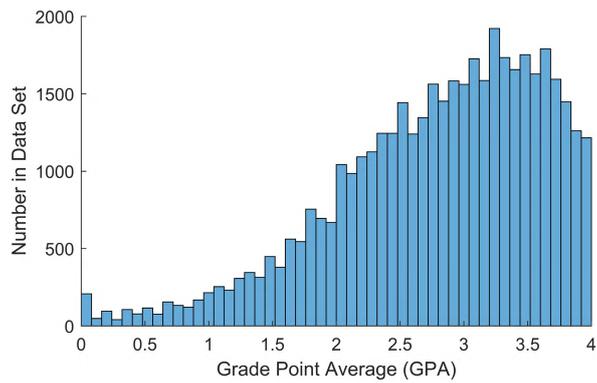

Figure 9: Histogram of student GPAs in the data set.

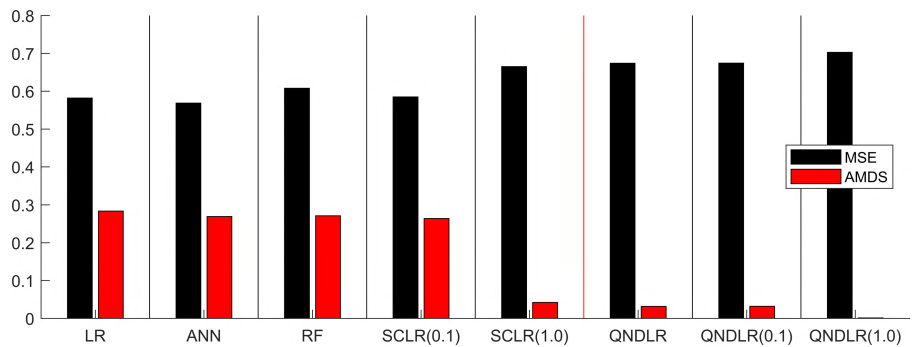

Figure 10: The mean squared error (MSE) and absolute mean discriminatory statistic (AMDS)—the absolute value of the mean sample discriminatory statistic—for the various regression algorithms.

Although in our prior work we developed Seldonian and quasi-Seldonian batch reinforcement learning algorithms [Thomas et al., 2015a; Thomas, 2015], we had not yet developed the Seldonian optimization framework and therefore did not think to generalize our methods to handle a broad class of behavioral constraint functions. That is, our past methods only allow for the single behavioral constraint: with high probability the *expected return* (performance) of the *policy* (solution) proposed by our algorithm should be at least some user-specified baseline. Still, the algorithm, *safe policy improvement* (SPI), that we proposed could be extended to provide the user with the ability to select behavioral constraints from within a broad class of possible constraints.

Rather than directly extend our previous work, here we will propose a new quasi-Seldonian algorithm that targets a different setting. We will take a nonstandard—but not uncommon [Kober and Peters, 2009; Sehnke et al., 2010; Theodorou et al., 2010; Stulp and Sigaud, 2012]—approach to reinforcement learning: we propose an algorithm that searches the space of distributions over policies rather than an algorithm that, like SPI, directly searches the space of policies. This decision will both simplify the use of *importance sampling* [Kahn and Marshall, 1953] by removing the product over time that usually appears when using importance sampling for reinforcement learning [Precup et al., 2000], and will make it easier for the user to understand what policies our algorithm could feasibly deploy.

Formally, let $\mathcal{P}$ be a set of (stochastic or deterministic) policies of interest for an episodic *Markov decision process* [Bertsekas and Tsitsiklis, 1996, MDP]. Intuitively, each $p \in \mathcal{P}$ denotes one way that the agent could make decisions, and might correspond to a neural network that takes as input the images coming from a video camera and produces as output the motor control signals for a robot [Levine et al., 2016]. Let $\mathcal{H}$ be the set of possible outcomes that could occur during one episode of an MDP—each $h \in \mathcal{H}$ is a sequence of states (observations), actions, and rewards that describes the agent's interactions with the environment during one episode. We refer to each $h \in \mathcal{H}$ as the *history* of an episode. Each policy, $p \in \mathcal{P}$, induces a distribution over the set of possible histories, $\mathcal{H}$. We write $H \sim p$ to denote that the history-valued random variable $H$ will be generated using the policy $p$. Let $r : \mathcal{H} \to \mathbb{R}$ be called the *return function*, and $r(h) \in \mathbb{R}$ be the *return* of the history, $h$, which is a measure of how "good" the history $h$ is, with larger values being preferable. Typically in reinforcement learning research the goal is to find an *optimal policy*, $p^\star$, which is a policy that maximizes the expected return:

$$p^\star \in \arg\max_{p \in \mathcal{P}} \mathbf{E}[r(H)|H \sim p].$$

We modify this standard formulation slightly—we will search for an optimal distribution over policies (within a feasible set of distributions over policies).

This allows for the application of our algorithm to control problems where the user has prior beliefs about which policies will perform well, and can therefore specify an initial distribution over (possibly deterministic) policies. Let $\mu_\theta$ be a distribution over $\mathcal{P}$ for all solutions, $\theta$. We write $P \sim \mu_\theta$ to denote that the policy-valued random variable $P$ will be sampled from $\mu_\theta$. So, the expected return when using a solution, $\theta$, can be written as $\mathbf{E}[r(H)|P \sim \mu_\theta, H \sim P]$.

We assume that $m \in \mathbb{N}_{>0}$ policies, $P_1, \ldots, P_m$, were sampled independently from some *behavior* distribution, $\mu^b$. These $m$ policies were then each used to generate one history, so that we have $m$ histories, $H_1, \ldots, H_m$. This *historical data*, $D$, will be the input to our algorithm, and is formally defined as the random variable: $D = \{(H_j, P_j)\}_{j=1}^m$. Notice that $D$ is a random variable because each $H_j$ and $P_j$ is a random variable. As before, let $\mathcal{D}$ be the set of all possible historical data sets, $D$. We can now state our goal as a SOP:

$$\arg\max_{a \in \mathcal{A}} \mathbf{E}[r(H)|P \sim \mu_{a(D)}, H \sim P] \tag{9}$$
$$\text{s.t. } \forall i \in \{1, \ldots, n\} \ \Pr(g_i(a(D)) \leq 0) \geq 1 - \delta_i,$$

where the expectation is taken with respect to the MDP faced by the user, and where $\mathcal{A}$ is the set of functions, $a \in \mathcal{A}$, where $a : \mathcal{D} \to \Theta$ (where $\Theta$ is the set of possible solutions, which we discuss later).

The algorithm that we will propose is likely not an optimal algorithmic solution (especially given that we do not know the MDP that the user will be faced with), although it is (quasi-)Seldonian. It will allow for a broad class of behavioral constraint functions, and provide the user with an expressive language with which to define them. Specifically, to specify the $i^\text{th}$ behavioral constraint function, $g_i$, the user must select an additional return function, $r_i : \mathcal{H} \to \mathbb{R}$, which is used to implicitly define the behavioral constraint:

$$g_i(\theta) \coloneqq \mathbf{E}[r_i(H)|P \sim \mu^b, H \sim P] - \mathbf{E}[r_i(H)|P \sim \mu_\theta, H \sim P].$$

That is, our algorithm must ensure that with probability at least $1 - \delta_i$ it will not change the distribution over policies to one that decreases the expected return, computed using the return function $r_i$. Notice that this means the user of the algorithm need not know which policies could cause desirable or undesirable behavior. For example, if the user can only recognize if a history, $h \in \mathcal{H}$ is an undesirable outcome, then he or she could define $r_i(h) = -1$ if $h$ is an undesirable outcome, and $r_i(h) = 0$ otherwise, to encode that with probability at least $1 - \delta_i$, the returned policy should cause undesirable behavior to occur with probably no more than it was under the behavior distribution.[2]

---

[2] It is straightforward to modify our algorithm to use the implicit definition $g_i(\theta) \coloneqq 1 - \beta_i - \mathbf{E}[r_i(H)|P \sim \mu_\theta, H \sim P]$, where $\beta_i$ is a user-specified constant. This modified form allows the user to require that the expected return be at least $1 - \beta_i$—a value that might not depend on the expected return under the behavior distribution. Using this modified form, selecting $r_i(h) = 0$ if $h$ is an undesirable outcome and $r_i(h) = 1$ otherwise would require that with probability at least $1 - \delta_i$ the policy will only be changed if the probability that the new policy produces an undesirable outcome is at most $\beta_i$.

We will focus on designing a quasi-Seldonian algorithm rather than a Seldonian algorithm. In our past work we found that Seldonian batch reinforcement learning algorithms often require large amounts of data, which although practical for big-data problems [Theocharous et al., 2015], often is not practical. By contrast, quasi-Seldonian methods using Student's $t$-test or bootstrap confidence bounds produced solutions given relatively small amounts of historical data, while still providing useful information about the probability that the behavioral constraints will be satisfied [Thomas, 2015]. We therefore chose to focus here on a quasi-Seldonian algorithm that uses confidence intervals based on Student's $t$-test. However, the approach that we specify here could be extended to use other confidence bounds, and proper use of a confidence bound like those produced by Hoeffding's inequality could produce a Seldonain algorithm.

To simplify the algorithm that we propose here, we assume that the set of solutions is small and finite—that $\theta \in \{1, \ldots, l\}$ for some small $l \in \mathbb{N}_{>0}$. This assumption allows us to avoid partitioning the training data into two sets—one of which is used to select a single candidate solution, and the other to determine whether this one solution satisfies the necessary probabilistic bounds. Instead, since $l$ is small, we will be able to use the union bound to ensure that our probabilistic bounds hold across all possible solutions simultaneously. This assumption means that the algorithm that we propose is mainly viable when the user has a few ideas about how the initial distribution, $\mu_b$, could be improved. Again, SPI [Thomas et al., 2015b] (extended to allow for more general behavioral constraints) is an example of how Seldonian and quasi-Seldonian algorithms might be designed without this assumption (it partitions the training data as described above).

Intuitively, the algorithm that we propose has two steps. Unlike the earlier (quasi-)Seldonian linear regression algorithms, which compute a single candidate solution from data, this algorithm assumes that $l$ candidate solutions have been provided. During the first step, it uses *high confidence off-policy policy evaluation* (HCOPE) methods, detailed in the works of Thomas et al. [2015a] and Thomas and Brunskill [2017], to test whether each of the $l$ solutions satisfy the behavioral constraints. Since these tests use the same data set when testing each of the potential policies, to avoid the problem of multiple comparisons [Miller, 2012], the bounds are each constructed to hold with probability at least $1 - \delta/l$, so that all hold simultaneously with probability at least $\delta$, by Boole's inequality. In the second step, the algorithm that we propose searches through the set of solutions that were deemed safe in the first step (if none are deemed safe it returns NO SOLUTION FOUND), and returns the single solution that it predicts will have the best performance.

We now describe these steps in more detail. First, for each of the $l$ possible solutions, each of the $n$ behavioral constraints, and each of the $m$ trajectories of historical data, we use *importance sampling* [Kahn and Marshall, 1953; Kahn, 1955] to construct an unbiased estimate, $\hat{\rho}_{i,j,k}$ (where $i \in \{1, \ldots, l\}$, $j \in \{1, \ldots, n\}$ and $k \in \{1, \ldots, m\}$), of $\mathbf{E}[r_j(H)|P \sim \mu_i, H \sim P]$, where $\mu_i$ is shorthand for $\mu_{\theta_i}$. In order for importance sampling to provide unbiased estimates, we require the assumption that $\operatorname{supp}(\mu_i) \subseteq \operatorname{supp}(\mu^b)$ for all $i \in \{1, \ldots, l\}$, which is standard in off-policy policy evaluation research [Precup et al., 2000; Thomas et al., 2015a; Jiang and Li, 2015]. However, if for some $i \in \{1,\ldots,l\}$, $\text{supp}(\mu_i) \neq \text{supp}(\mu^b)$, we can leverage our knowledge of $\mu^b$ and $\mu_i$ to improve the ordinary importance sampling estimates [Thomas and Brunskill, 2017]. In our algorithm we use this modified importance sampling estimator to construct each $\hat{\rho}_{i,j,k}$ (the modified importance sampling estimator may return fewer than $m$ estimators for each $i, j$ pair, and so $k \in \{1,\ldots,m'\}$ for some $m' \leq m$).

Once $\hat{\rho}_{i,j,k}$ has been computed for all values of $i, j$, and $k$, we use Student's $t$-test to get high confidence approximate lower-bounds on $\mathbf{E}[r_j(H)|P \sim \mu_i, H \sim P\mu]$ for all $i \in \{1,\ldots,l\}$ and $j \in \{1,\ldots,n\}$. We also use Student's $t$-test to compute high confidence approximate upper-bounds, $\beta_j$, on $\mathbf{E}[r_j(H)|P \sim \mu^b, H \sim P\mu]$ for all $j \in \{1,\ldots,n\}$. Importantly, here we apply Student's $t$-test using $\delta_j/(l+1)$ so that the high-probability approximate upper bound and lower bounds for all solutions all hold simultaneously. Next, our algorithm checks which solutions—which values of $i \in \{1,\ldots,l\}$—produced high confidence approximate lower bounds that were all above their respective baseline values, $\beta_j$. Those solutions that did are deemed *safe*, because our algorithm could return any of these solutions and it would be quasi-Seldonian. If no solutions are deemed safe, then our algorithm returns NO SOLUTION FOUND (NSF).

If at least one solution is deemed safe, then our algorithm searches for the solution, within the safe set of solutions, that it predicts will produce the largest expected return on the MDP. The predictions of expected returns are once again constructed using the modified form of importance sampling. Pseudocode for our algorithm is provided in Algorithm 12 (for simplicity, this pseudocode assumes that each distribution over policies is a probability density function, and uses Riemann integrals).

## 7.1 On the Ease of Using Our Quasi-Seldonian Reinforcement Learning Algorithm

Algorithms designed using our framework should be easier for a user to deploy responsibly than algorithms designed using the standard approach. We have already shown how difficult it can be to include probabilistic constraints on the behavior of an algorithm designed using the standard approach. Here we point out that our quasi-Seldonian reinforcement learning algorithm is a strong example of a new type of algorithm that makes it easy for the user to place probabilistic constraints on the algorithm's behavior.

In order to ensure with high probability that undesirable behavior does not occur, when using our new algorithm the user does *not* need to perform additional data analysis, does *not* need to have training in statistics and data mining, and does *not* need to have detailed knowledge of the problem at hand (such as knowledge about the true transition function or reward function of a Markov decision process). Instead, the user of the algorithm need only be able to recognize when undesirable behavior has occurred: the user must be able to define $r_i(H)$ to be a measure of how much undesirable behavior occurred in the outcome $H$ (as discussed previously). Furthermore, other than the auxiliary reward functions, $r_i$, and their corresponding confidence levels, $\delta_i$, our algorithm has no additional hyper-parameters that the user must tune. So, our new algorithm can easily be applied by an unskilled user to a variety of batch reinforcement learning problems where the user is able to define what constitutes undesirable behavior, but may not know which policies cause undesirable behavior.

## 7.2 Application of Quasi-Seldonian Reinforcement Learning Algorithm to Diabetes Treatment

We applied the quasi-Seldonian reinforcement learning algorithm presented in Algorithm 12 to the problem of providing personalized improvements to treatment policies for type 1 diabetes (Diabetes mellitus type 1). At a high level, the body of a person with type 1 diabetes (hereafter, simply *diabetes*) does not produce sufficient *insulin*—a hormone that promotes the absorption of *glucose* (a type of sugar) from the blood into some types of cells. As a result, people with untreated diabetes tend to have high blood glucose levels—a condition called *hyperglycemia*, which can have severe consequences [World Health Organization, 2016]. One treatment for diabetes is the injection of insulin into the blood several times each day. If too much insulin is injected, blood glucose levels can become too low—a condition called *hypoglycemia*. Although hyperglycemia is undesirable, it is minor in comparison to instances of severe hypoglycemia, which can triple the five-year mortality rate for a person with diabetes [McCoy et al., 2012].

The amount of insulin (measured using *insulin units*) that a person with diabetes is instructed to inject prior to eating a meal is given by:

$$\text{injection} = \frac{\text{blood glucose} - \text{target blood glucose}}{CF} + \frac{\text{meal size}}{CR},$$

where "blood glucose" is an estimate of the person's current blood glucose (measured from a blood sample and using mg/dL), "target blood glucose" is the desired blood glucose (also measured using mg/dL), which is typically specified by a diabetologist, "meal size" is an estimate of the size of the meal to be eaten (measured in grams of carbohydrates), and $CR$ and $CF$ are two real-valued parameters that, for each person, must be tuned to make the treatment policy effective. Usually a diabetologist will define the target blood glucose and make an educated guess as to which values of $CR$ and $CF$ will work well for a person. These values are adjusted by the diabetologist during follow-up visits every 3–6 months. However, in some instances (particularly in countries with low Human Development Indexes) it is not possible for a person to see a diabetologist this frequently.

Researchers have therefore proposed the use of reinforcement learning algorithms to automatically tune $CR$ and $CF$ when follow-up visits to a diabetologist are not an option [Bastani, 2014]. In this context, $CR$ and $CF$ are the parameters of a parameterized policy for an MDP where actions correspond to recommended injection sizes and the state is a complete description of the person at the current time. The parameterized policy uses *function approximation*—it depends only

**Algorithm 12:** Quasi-Seldonian Reinforcement Learning Algorithm

**1 Input: 1)** Data set $D = \{(H_j, P_j)\}_{j=1}^m$, **2)** behavior distribution, $\mu^b$, **3)** MDP return function, $r : \mathcal{H} \to \mathbb{R}$, **4)** $n$ behavioral constraint return functions, $r_1, \ldots, r_n$, where each $r_i : \mathcal{H} \to \mathbb{R}$, **5)** $n$ behavioral constraint confidence levels, $\delta_1, \ldots, \delta_n$, where each $\delta_i \in (0, 1)$, and **6)** $l$ candidate distributions over policies, $\mu_1, \ldots, \mu_l$, where $\text{supp}(\mu_i) \subseteq \text{supp}(\mu^b)$ for all $i \in \{1, \ldots, l\}$.

**2 Assumptions:** This algorithm assumes that the sum of roughly $m$ i.i.d. random variables (the importance weighted returns) is normally distributed.

```
/* Upper bound the expected returns if μ^b and each r_j were
   to be used.                                               */
```

**3 for** $j = 1$ *to* $n$ **do**

**4** $\quad \beta_j \leftarrow \frac{1}{m} \sum_{k=1}^m r_j(H_k) + \frac{1}{\sqrt{m}} \text{stddev}([r_j(H_1), r_j(H_2), \ldots, r_j(H_m)]) \text{tinv}(1 - \delta_j/(l+1), m-1)$;

```
/* Determine which solutions, θ ∈ {1,...,l} are safe.  Loop
   over each solution and test whether it should be removed
   from safe.                                               */
```

**5** safe $\leftarrow \{1, 2, \ldots, l\}$;  // Set of integers.

**6 for** $i = 1$ *to* $l$ **do**

**7** $\quad c_i \leftarrow \int_{\text{supp}(\mu_i)} \mu^b(p) \, dp$;  // Scalar.

```
    /* Check whether the i^th solution satisfies the j^th
       behavioral constraint.                               */
```

**8** $\quad$ **for** $j = 1$ *to* $n$ **do**

```
        /* Load ρ̂ with all of the importance weighted returns.
                                                            */
```

**9** $\quad\quad$ Create empty list, $\hat{\rho}$, of floating point numbers;

**10** $\quad\quad$ **for** $k = 1$ *to* $m$ **do**

**11** $\quad\quad\quad$ **if** $\mu_i(P_k) \neq 0$ **then** append $c_i \frac{\mu_i(P_k)}{\mu^b(P_k)} r_j(H_k)$ to the list $\hat{\rho}$;

```
        /* Check whether the lower bound is less than the
           baseline, β_j, and if it is, then mark the i^th
           solution as unsafe.                              */
```

**12** $\quad\quad$ **if** $\hat{\rho} = \emptyset$ **or** $\text{mean}(\hat{\rho}) - \frac{\text{stddev}(\hat{\rho})}{\sqrt{\text{length}(\hat{\rho})}} \text{tinv}(1 - \delta_j/(l+1), \text{length}(\hat{\rho}) - 1) < \beta_j$
$\quad\quad$ **then** safe $\leftarrow$ safe $\setminus \{i\}$;

```
/* The set "safe" now holds the set of safe solutions.  If
   it is empty, return NSF. If it is not empty, search for
   the solution predicted to maximize expected return.    */
```

**13 if** safe $= \emptyset$ **then return** NO SOLUTION FOUND (NSF);

**14 for** $idx = 1$ *to* $\text{length}(\text{safe})$ **do**

**15** $\quad i \leftarrow \text{safe}[idx]$;  // Scalar integer.

**16** $\quad$ curPerf $\leftarrow \left( c_i \sum_{k=1}^m \frac{\mu_i(P_k)}{\mu^b(P_k)} r(H_k) \right) / \left( \sum_{k=1}^m \mathbf{1}_{(\mu_i(P_k) \neq 0)} \right)$;  // Scalar.

**17** $\quad$ **if** $idx = 1$ **or** $\text{curPerf} > \text{bestPerf}$ **then**

**18** $\quad\quad$ bestPerf $\leftarrow$ curPerf;  // Scalar.

**19** $\quad\quad$ bestIdx $\leftarrow$ idx;  // Scalar integer.

**20 return** safe[bestIdx];

on an observation (the blood glucose estimate acquired from a blood sample) that depends on the current state of the person.[3]

Intuitively, the reward function for the MDP should penalize deviations of blood glucose from optimum levels (with larger penalties for blood glucose levels that are too low). Precisely defining a reward function of this sort is difficult because there are two conflicting goals: **1)** keep blood glucose levels from becoming too high, and **2)** keep blood glucose levels from becoming too low. Because hypoglycemia can have much more severe consequences than hyperglycemia, the goal is typically to minimize the time that a person is hyperglycemic subject to the constraint that hypoglycemia never occurs. In practice, however, hypoglycemia cannot be completely precluded [McCoy et al., 2012]. This means that the reward function must be selected to trade-off between hyperglycemia and hypoglycemia—a problem described in detail by Bastani [2014, Section 1.3.6].

As discussed previously when describing the limitations of soft constraints, properly selecting the reward function to balance the two conflicting objectives is difficult and requires detailed knowledge of the problem at hand. Later we will describe how this problem can be mitigated by using a Seldonian algorithm. For now, we begin by defining the MDP to use a reward function that is similar to that proposed by Bastani [2014]: the reward associated with a blood glucose measurement of $bg$ (in mg/dL) is given by

$$r(bg) := \begin{cases} -\frac{(bg'-6)^2}{5} & \text{if } bg' < 6 \\ -\frac{(bg'-6)^2}{10} & \text{if } bg' \geq 6, \end{cases}$$

where $bg' := bg/18.018018$ is an estimate of $bg$ in mmol/L rather than mg/dL.[4] We also define an auxiliary reward function, $r_1$, that we will use later to create a behavioral constraint. Unlike the MDP reward function, $r$, the auxiliary reward function, only penalizes low blood glucose levels:

$$r_1(bg) := \begin{cases} -\frac{(bg'-6)^2}{5} & \text{if } bg' < 6 \\ 0 & \text{if } bg' \geq 6. \end{cases} \tag{10}$$

For these experiments, we used an *in-silico* (simulated) subject, simulated using a newer version of the simulator used by Bastani [2014], which is called the *type 1 diabetes metabolic simulator* (T1DMS) [Dalla Man et al., 2014]. T1DMS is a metabolic simulator that has been approved by the US Food and Drug Administration as a substitute for animal trials in pre-clinical testing of treatment policies for type 1 diabetes. We selected subject adult#003 within T1DMS as a representative subject, and focus on this single subject since our goal is to show that undesirable behavior can occur when using algorithms designed using the standard approach, and that a Seldonian algorithm can find policy improvements using a realistic amount of data (and existential arguments require only a single example). Following the experimental setup proposed by Bastani [2014], the subject is provided with three meals of randomized sizes at randomized times. This combined with the inherent stochasticity of the T1DMS model means that applying the same $CR$ and $CF$ values for two different days can result in different outcomes.

---

[3]Alternatively, this problem could be modeled as a *partially observable Markov decision process* (POMDP). However, since we focus on *state-free policies* [Kaelbling et al., 1996, Section 7], the MDP framework with function approximation is sufficient.

[4]We primarily use mg/dL, since it is currently preferred in the United States and is the default unit used by T1DMS. We define the reward function using mmol/L to show its similarity to the reward function used by Bastani [2014], who used mmol/L.

First, we estimated the expected returns (undiscounted sums of rewards) that results from using either the MDP reward function or the auxiliary reward function, and using various values of $CR$ and $CF$ within a reasonable range, $CR \in [5, 50]$ and $CF \in [1, 31]$. These estimates are depicted in Figure 11. Notice that, when $CR$ is too small, hypoglycemia occurs often—the mean returns using the auxiliary reward function or MDP reward function are both large negative values. When $CR$ is too large, instances of hypoglycemia decrease (the mean return using the auxiliary reward function plateaus at zero), but instances of hyperglycemia increase (the mean return using the MDP reward function decreases). Thus, the goal of an agent selecting values for $CR$ and $CF$ is to find values that lie along the ridge of the objective function that occurs around $CR \approx 10$, erring towards values of $CR$ that are too large.

Notice that these plots are typically *not* available to the diabetologist selecting initial values for $CR$ and $CF$, nor are they available to an algorithm that might try to improve upon a diabetologist's initial educated guess as to what values of $CR$ and $CF$ will work for the subject. Furthermore, it is difficult for any agent (a diabetologist or reinforcement learning agent) to accurately estimate these plots from data—each plot was the result of 44,000 simulated days. To visualize the difficulty of estimating the expected return for a single $CR$ and $CF$ pair, consider Figure 12, which shows the result of applying two different $CR$ and $CF$ pairs—one that lies just beyond the ridge, and which results in frequent hypoglycemic episodes, and the other which is near-optimal and lies near the top of the ridge. Notice the high variance of the returns relative to the differences between the expected returns using different values of $CR$ and $CF$ (c.f. Figure 11). This high variance means that it is difficult to determine from small amounts of data which of these two $CR$ and $CF$ pairs is better, let alone search the uncountably infinite space of $CR$ and $CF$ pairs for their optimal settings. Thus, one might expect that the high confidence guarantees required of (quasi-)Seldonian algorithms might be impractical. We will show that, even for this challenging problem, quasi-Seldonian algorithms can be effective.

In the preliminary study performed by Bastani [2014], the parameterized policy was stochastic—the amount of insulin that it proposed injecting included (bounded) random noise. This noise results in *exploration*, and is often necessary for learning to proceed [Sutton and Barto, 1998]. However, it makes it difficult for one to predict which sequences of injections the policy might propose. To avoid this problem, we modify the problem setting used by Bastani [2014] so that the diabetologist initially proposes a *range* of values for $CR$ and $CF$. The values for $CR$ and $CF$ that are used each day are then sampled from the uniform distribution over this range. This provides an intuitive means for the diabetologist to specify acceptable exploration.

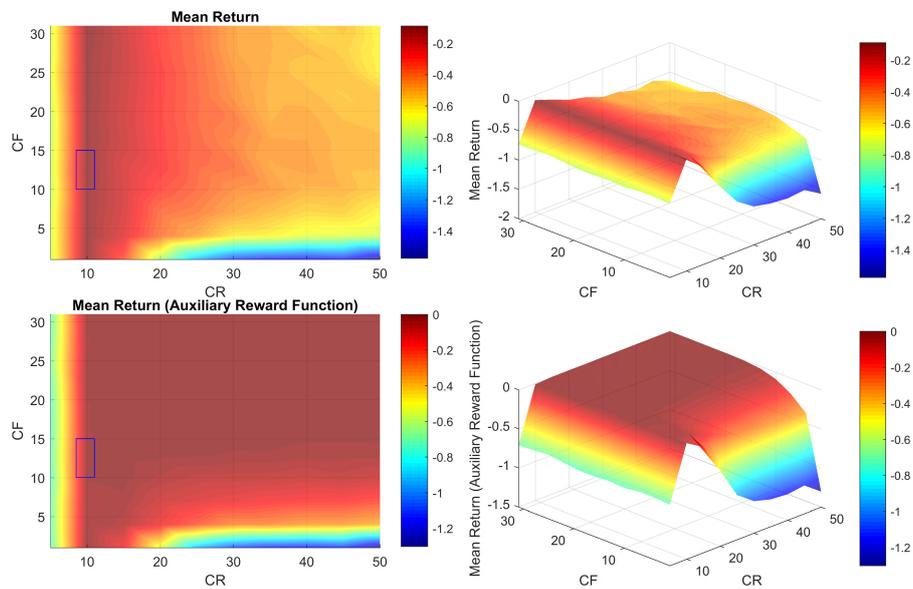

Figure 11: The mean returns using the reward function of the MDP, which penalizes both hyperglycemia and hypoglycemia (top row), and the mean returns using the auxiliary reward function, $r_1$, which only penalizes hypoglycemia (bottom row). The blue box depicts an initial range of $CR$ and $CF$ values that might be selected by a diabetologist. The values of $CR$ were tested at intervals of 5 while the values of $CF$ were tested at intervals of 3. Each $CR$ and $CF$ pair was evaluated using Monte Carlo sampling—500 days worth of data. Values between the grid of sampled points are interpolated using linear interpolation.

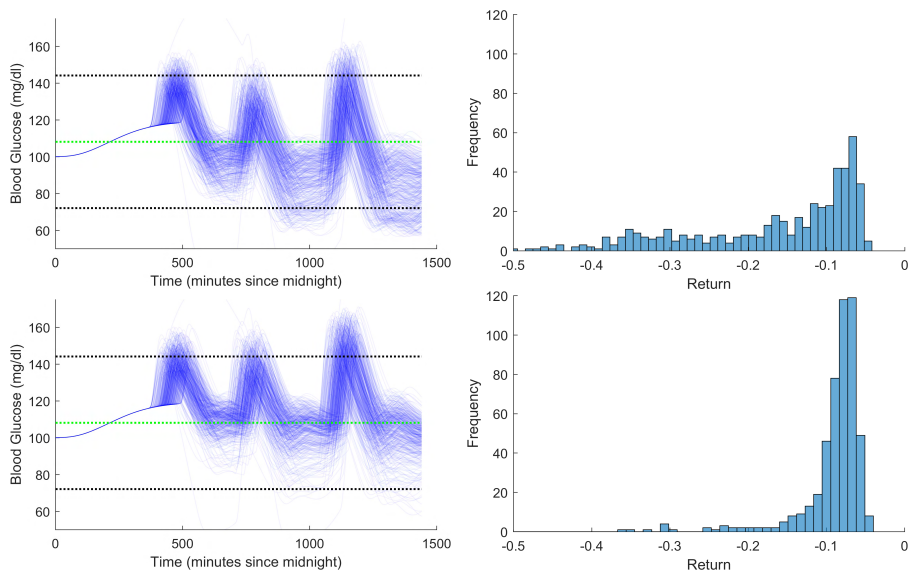

Figure 12: Examples of 300 different days using $CR = 8.5$, $CF = 10$ (top row) and $CR = 10$, $CF = 12.5$ (bottom row). The former results in frequent hypoglycemia, while the latter is a near optimal policy. The plots on the left show the blood glucose throughout the day, with the black lines denoting a desirable range of blood glucose levels [Bastani, 2014], and the green line denoting optimal levels. The three increases in blood glucose correspond to the times when the subject eats breakfast, lunch, and dinner. The plots on the right show histograms of the resulting returns (sum of rewards) computed using the MDP reward function (returns less than $-0.5$ occur but are not shown).

Due to a lack of data when the first $CR$ and $CF$ values are selected by the diabetologist, they are not always optimal. For our *in silico* subject, we assume that the diabetologist specifies the initial range, $CR \in [8.5, 11]$ and $CF \in [10, 15]$—the blue box in Figure 11. That is, the initial distribution over policies, $\mu^b$, is the uniform continuous distribution over this interval. Figure 13 provides a zoomed in view of the objective function over this range. Notice that this interval contains near optimal policies, where $CR \approx 10$, as well as some undesirable policies (e.g., when $CR \approx 8.5$). Furthermore, notice that we observe the same trend as Bastani [2014]: when $CR$ is selected properly, performance is robust to the value of $CF$.

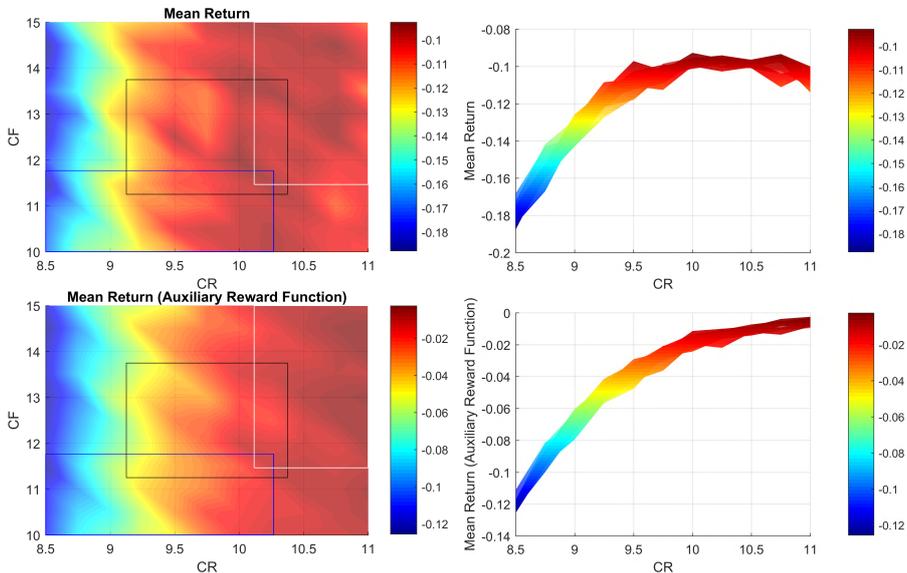

Figure 13: A zoomed in view of Figure 11. The boundaries of these plots are the range of $CR$ and $CF$ used by the initial distribution over policy parameters, $\mu^b$. The plots on the right show a side view, which shows the difference between the reward function of the MDP (which penalizes both hyperglycemia and hypoglycemia) and the auxiliary reward function, $r_1$, which only punishes hypoglycemia. For all of these plots, $CR$ was varied by increments of 0.25 and $CF$ was varied by increments of 0.5, each $CR$ and $CF$ pair was evaluated using Monte Carlo sampling with 500 days of data, and values between sampled $CR$ and $CF$ values are interpolated using linear interpolation. The blue, black, and white boxes in the plots on the left are discussed later in the text.

We consider the problem of applying a batch reinforcement learning agent that takes $m \in \mathbb{N}_{>0}$ days of data collected from values of $CR$ and $CF$ sampled from $\mu^b$ (each pair of policy parameters is used for one day), and which produces as output a new distribution over policies. Ideally, the algorithm should be able to propose a new policy (or distribution over policies) using around 6 months of data—$m \approx 180$. Applying a reinforcement learning algorithm designed using the standard approach, like that proposed by Bastani [2014] presents at least two serious problems. First, as discussed previously when describing the limitations of soft constraints, properly selecting the reward function to balance the two conflicting objectives is difficult and requires detailed knowledge of the problem at hand. The reward function that we have chosen is intuitive, but we have no guarantees (or meaningful probabilistic expectations) about whether or not it will result in policies that increase or decrease the prevalence of hypoglycemia relative to $\mu^b$ (the initial policy distribution specified by the diabetologost).

The second serious drawback of reinforcement learning algorithms designed using the standard approach is that they do not provide meaningful guarantees about their performance. If the hyperparameters of the algorithm (like the step size, the granularity of function approximators used to estimate the value function, or the exploration rate) are not tuned properly, then the reinforcement learning algorithm might often produce policies (values for $CR$ and $CF$) that are significantly worse than the initial policy proposed by the diabetologist. Furthermore, even if the hyperparameters are properly tuned, then during the search for better values for $CR$ and $CF$, the algorithm might still deploy values of $CR$ and $CF$ that increase the frequency of hypoglycemia.

We propose using the quasi-Seldonian algorithm presented in Algorithm 12 to remedy these problems. Intuitively, we would like a guarantee that, if the reinforcement learning algorithm changes the distribution over policies proposed by the diabetologist, then it will not increase the prevalence of hypoglycemia. In practice, this is not possible—the observed data could be an unlikely sample that does not reflect the real world, and so any algorithm might sometimes change the policy so that the prevalence of hypoglycemia increases. So, instead, we require the reinforcement learning algorithm to guarantee that the probability that it will change the distribution over policies to one that increases the prevalence of hypoglycemia will be small. Furthermore, we might want to ensure that the reinforcement learning algorithm will never deploy values of $CR$ and $CF$ outside the range specified by the diabetologist.

These desired properties of a batch policy search algorithm are captured by the Seldonian optimization problem presented in (9), where the set of algorithms, $\mathcal{A}$, is restricted to only contain algorithms that will never return values of $CR$ and $CF$ outside $CR \in [8.5, 11]$ and $CF \in [10, 15]$, and where there is a single behavioral constraint ($n = 1$), with auxiliary reward function $r_1$, as defined in (10). That is, a (quasi-)Seldonian algorithm must ensure that with high probability the prevalence of hypoglycemia (measured using $r_1$) will not increase, and should try to maximize the expected return (using the MDP reward function) otherwise. Because safety is critical, we selected a small constraint confidence level: $\delta_1 = 0.05$.

Algorithm 12 is a viable quasi-Seldonian algorithm for this SOP, which allows for the specification of a set of possible distributions over policies that will be considered. We selected 27 such distributions, each of which is a uniform distribution over ranges for $CR$ and $CF$ that are subsets of the support of $\mu^b$. Furthermore, each of the 27 distributions contains $1/4$ the support of $\mu^b$. The 27 distributions were generated by an automatic tiling scheme to sample various ratios of the range of $CR$ to the range of $CF$, and positionings within the support of $\mu^b$. Examples of the ranges of three of these 27 distributions are provided in Figure 13 as blue, black, and white boxes.

We modified Algorithm 12 to include a model-based control variate in the importance sampling estimate [Hammersley, 1960; Jiang and Li, 2015; Thomas and Brunskill, 2016]. The approximate model used to construct the control variate was a different setting of T1DMS that produces similar responses to adult#003, but which is not the same. Importantly, the optimal values of $CR$ and $CF$ under the approximate model cause frequent episodes of hypoglycemia—the model cannot be trusted to select values for $CR$ and $CF$. However, the approximate model does provide an effective control variate, which decreases the variance of importance sampling estimates. Furthermore, we observed similar (although not quite as strong) results without the use of this model-based control variate.

Figure 14 shows the result of deploying Algorithm 12, as well as a non-Seldonian algorithm. The non-Seldonian algorithm that we selected uses importance sampling (with the model-based control variate) to estimate the expected return if each of the 27 possible policy distributions were used, and returns the distribution that it predicts will result in the largest return (in terms of the MDP reward function). This corresponds to Algorithm 12, but without any behavioral constraints. Notice that the quasi-Seldonian algorithm is able to return solutions (other than NSF, in which case the diabetologist's policy is not changed) given very little data—as little as one month of data. Given 6 months of data (roughly 180 days), the quasi-Seldonian algorithm almost always returns a new distribution for $CR$ and $CF$ that is different from the initial distribution. While the Seldonian algorithm sometimes does not return a new policy distribution, the algorithm designed using the standard approach always does. However, this comes at a cost—the algorithm designed using the standard approach often deploys policies that result in increased prevalence of hypoglycemia. By contrast, across all trials, the quasi-Seldonian algorithm never returned a distribution over policies that increased the prevalence of hypoglycemia.[5] That is, the algorithm designed using the standard approach would not be safe to deploy for adult#003, but the quasi-Seldonian algorithm would be.

Notice also that Figure 14 shows that the costs associated with providing high probability safety guarantees are relatively minor. Specifically, notice that given roughly 75 days of data, our empirical results suggest that with probability at least 95%, the non-Seldonian algorithm returns policies that do not increase the prevalence of hypoglycemia (when applied to adult#003). Similarly, the quasi-Seldonian algorithm begins returning solutions frequently given roughly 100–150 days of data. This slight lag between when the non-Seldonian algorithm begins to act in a safe manner and when our quasi-Seldonian algorithm begins returning solutions (when it is able to automatically determine that returning solutions would be safe) is the cost associated with requiring a safety guarantee. If the quasi-Seldonian algorithm was not data efficient, then it would not begin returning solutions frequently until long after the non-Seldonian algorithm tended to return solutions that do not increase the prevalence of hypoglycemia.

---

[5]Without the control variate, the quasi-Seldonian algorithm sometimes did return distributions over policies that increased the prevalence of hypoglycemia, but still with probability well below 0.05.

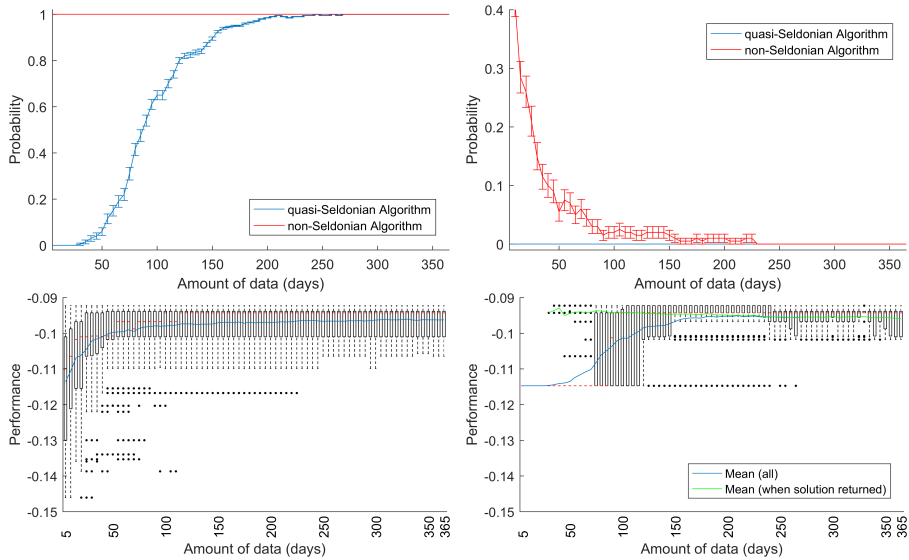

Figure 14: **Top Left:** The probability that a solution other than NO SOLUTION FOUND (NSF) is returned. **Top Right:** The probability that a distribution over policies was returned with lower expected return than the initial distribution over policies, using the auxiliary reward function $r_1$. That is, the probability that a solution was returned which increases the prevalence of hypoglycemia. **Bottom Left:** For different numbers of days of data (in intervals of 5), a box-plot of the distribution of expected returns of the solutions produced by the algorithm designed using the standard approach. Outliers are shown as black dots, and the blue line is the mean return. **Bottom Right:** The same as the bottom left plot, but for the quasi-Seldonian algorithm. The blue line is the mean return, where the initial distribution over policies is used if the algorithm returns NSF, and the green line is the mean return given that a solution other than NSF is returned.

In this study the quasi-Seldonian algorithm was provided with a range of admissible values for $CR$ and $CF$ for adult#003, i.e., $CR \in [8.5, 11]$ and $CF \in [10, 15]$, which is a subset of the reasonable ranges for these parameters ($CR \in [5, 50]$ and $CF \in [1, 31]$), as depicted in Figure 11. This admissible set of values for $CR$ and $CF$ is a hard constraint, i.e., the specification of a feasible set. We provided our Seldonian algorithm with this feasible set, which places limitations of the solutions that it can return, because in this application it is reasonable to assume that a diabetologist could provide this initial range of reasonable parameters (and may also appreciate the guarantee that only parameters from within this range will be used).

This raises the question: was the safety of our quasi-Seldonian algorithm due to our providing it with this feasible set? We contend that the answer to this question is "no." The feasible set that we used contains dangerous policies—policies that cause frequent instances of hypoglycemia for adult#003. So, our example is one where the diabetologist selects a region of policy space (range for $CR$ and $CF$) that is dangerous and suboptimal. In cases where the diabetologist correctly identifies optimal settings for $CR$ and $CF$, there is no need for tuning, and most algorithms would be safe—the crucial setting is that where the initial range contains suboptimal policies and should be adjusted. Given that the initial range of values for $CR$ and $CF$ that we used includes dangerous policies, it is therefore important that an algorithm that automatically adjusts the treatment policy does not deploy a new treatment policy (within the window specified by the diabetologist) that increases the prevalence of hypoglycemia. Figure 14 shows that this could happen: the non-Seldonian algorithm returned solutions that frequently increased the prevalence of hypoglycemia *even though it too was restricted to only return solutions within the feasible set specified by the diabetologist.*

## 8  Other Seldonian Algorithms

There has been growing interest in ensuring that machine learning algorithms are safe to use—see for example the works of Amodei et al. [2016], Russell [2016], and Zilberstein [2015]. Given the practical nature of the guarantees provided by Seldonian algorithms, it would therefore be surprising if there were not already some algorithms that can be viewed as (quasi-)Seldonian algorithms for specific Seldonian optimization problems. Here we discuss some existing (quasi-)Seldonian algorithms and the problems that they solve. However, to the best of our knowledge, the framework of Seldonian optimization problems has not been proposed as a general problem formulation for machine learning, nor has its benefits been thoroughly discussed previously.

One example of both Seldonian and quasi-Seldonian algorithms is our prior work to create reinforcement learning algorithms for digital marketing applications [Thomas et al., 2015a,b; Thomas, 2015]. These algorithms observe a vector of features that describe the information that is known about a person visiting a webpage, and decide which advertisement, or which type of advertisement, to display on the webpage. The deployment of a policy that is worse than existing policies would result in fewer clicks on the advertisements, which in turn could be costly both in terms of lost advertisement revenue and lost customers for a digital marketing product.

In the context of digital marketing, we proposed Seldonian and quasi-Seldonian batch reinforcement learning algorithms that guarantee that with probability at least $1 - \delta$, their performance (in terms of the expected return of the MDP) will be at least some constant, $\rho_-$, where the user of the algorithm is free to select $\delta$ and $\rho_-$. Although this prior work was a stepping stone towards this work, it lacked several important features. First, we did not allow for any other behavioral constraints—the only constraint that we considered was ensuring that the expected return was increased with high probability. Second, we only considered the reinforcement learning setting—we did not observe that behavioral constraints could be important for other branches of machine learning.

Others have considered the problem of guaranteeing policy improvement with high confidence in the context of determining how the complexity of (approximately) solving MDPs grows with different parameters of the MDP, like the number of possible states [Kakade, 2001]. Others have also considered constructing confidence intervals around performance estimates constructed using *counterfactual reasoning* [Bottou et al., 2013]—an approach similar to, and preceding, our own prior work [Thomas et al., 2015a]. Like our prior work, these examples consider only one special case of SOPs: SOPs that model reinforcement learning (or bandit) problems and contain the single behavioral constraint that performance (in terms of the standard objective function) should be increased with high probability. This single behavioral constraint is common and not unique to reinforcement learning either: the problem of bounding the generalization error of a supervised learning algorithm has been well studied [Abu-Mostafa et al., 2012].

Another example of a (quasi-)Seldonian algorithm is the algorithm presented by Berkenkamp et al. [2016], which was developed in parallel with, and independently of, this work. Berkenkamp et al. [2016] propose a special case of the Seldonian optimization problem framework, wherein the goal is to tune the hyperparameters of a control algorithm to ensure that, with high probability, the controller will avoid unsafe regions of state space. They then propose a quasi-Seldonian algorithm that uses Gaussian processes to approximate both the objective function and behavioral constraint functions, and which then acts conservatively with respect to the confidence bounds produced by the Gaussian process. This example provides yet another example showing the viability of Seldonian optimization problems. However, Berkenkamp et al. [2016] do not discuss how the problem framework can be generalized to other machine learning applications.

Furthermore, one might view many algorithms as solutions to specific Seldonian optimization problems. For example, many of the state-avoidant algorithms that we discussed in §4.4—e.g., the algorithms proposed by Akametalu et al. [2014]—ensure that with high probability they will not allow a system to enter a pre-specified undesirable state. Thus, these algorithms are (quasi-)Seldonian algorithms for a SOP that includes the behavioral constraint that, with high probability, the agent will never enter an undesirable state (and which assumes that at least some prior knowledge about the system dynamics is available). Moreover, as the Seldonian optimization framework subsumes the standard machine learning optimization framework, most existing machine learning algorithms can be viewed as instances of (quasi-)Seldonian algorithms.

# Acknowledgements

The research reported here was supported in part by a NSF CAREER grant and a grant from Microsoft. The opinions expressed are those of the authors and do not represent views of the NSF or Microsoft.

# A  Derivation of Minimum Mean Squared Error Estimator for Illustrative Example

In this appendix we show that the minimum mean squared error estimator of $Y$ given $X$ in our illustrative example proposed in §4.1 is $\frac{2}{3}X$. Here we do not limit our search of estimators to linear functions—$\frac{2}{3}X$ has the lowest mean squared error of all possible estimators that use $X$ (and only $X$) to predict $Y$. To show this result, we derive an expression for the minimum mean squared error estimate of $Y$ given that $X = x$.

We begin by writing an expression for the *mean squared error* (MSE) of any estimate, $\hat{y} \in \mathbb{R}$, given that $X = x$.

$$\text{MSE}(\hat{y}) \coloneqq \int_{-\infty}^{\infty} \Pr(Y = y | X = x)(\hat{y} - y)^2 \mathrm{d}y,$$

where, in this section of the appendix only, we abuse notation and write Pr to denote probability densities rather than probabilities. To find the value of $\hat{y}$ that minimizes MSE we find the critical points of MSE:

$$\begin{aligned} 0 =& \frac{\partial}{\partial \hat{y}} \text{MSE}(\hat{y}) \\ =& \frac{\partial}{\partial \hat{y}} \int_{-\infty}^{\infty} \Pr(Y = y | X = x)(\hat{y} - y)^2 \, \mathrm{d}y \\ =& 2 \int_{-\infty}^{\infty} \Pr(Y = y | X = x)(\hat{y} - y) \, \mathrm{d}y. \end{aligned} \quad (11)$$

By Bayes theorem and marginalizing over $T$, we have that:

$$\Pr(Y=y|X=x) = \frac{\Pr(X=x|Y=y)\Pr(Y=y)}{\Pr(X=x)}$$
$$= \frac{\Pr(X=x|Y=y)\sum_{t=0}^{1}\Pr(T=t)\Pr(Y=y|T=t)}{\Pr(X=x)}.$$

Thus, continuing from (11) we have that:

$$0 = 2\int_{-\infty}^{\infty} \frac{\Pr(X=x|Y=y)\sum_{t=0}^{1}\Pr(T=t)\Pr(Y=y|T=t)}{\Pr(X=x)}(\hat{y}-y)\,dy$$

$$= 2\int_{-\infty}^{\infty} \underbrace{\frac{1}{2\pi}e^{-\frac{(x-y)^2}{2}}}_{\Pr(X=x|Y=y)}\left(\underbrace{0.5}_{\Pr(T=0)}\underbrace{\frac{1}{2\pi}e^{\frac{-(y-1)^2}{2}}}_{\Pr(Y=y|T=0)} + \underbrace{0.5}_{\Pr(T=1)}\underbrace{\frac{1}{2\pi}e^{\frac{-(y+1)^2}{2}}}_{\Pr(Y=y|T=1)}\right)\frac{(\hat{y}-y)}{\Pr(X=x)}\,dy$$

$$= \frac{1}{4\pi^2\Pr(X=x)}\int_{-\infty}^{\infty} e^{-\frac{2(x-y)^2+y^2}{4}}(\hat{y}-y)\,dy.$$

Thus,

$$\left(\int_{-\infty}^{\infty} e^{-\frac{2(x-y)^2+y^2}{4}}\,dy\right)\hat{y} = \int_{-\infty}^{\infty} e^{-\frac{2(x-y)^2+y^2}{4}}y\,dy$$

$$\left(2\sqrt{\frac{\pi}{3}}e^{-\frac{x^2}{6}}\right)\hat{y} = \left(\frac{4}{3}\sqrt{\frac{\pi}{3}}e^{\frac{-x^2}{6}}\right)x$$

$$\hat{y} = \frac{2}{3}x.$$

It is straightforward to verify that this unique critical point is a global minimum of $\text{MSE}(\hat{y})$ (as opposed to a saddle point or maximum).